\documentclass{article}

\usepackage{microtype}
\usepackage{graphicx}
\usepackage{subcaption}
\usepackage{booktabs} %

\usepackage{hyperref}

\usepackage[preprint]{icml2026}
\usepackage{amsmath}
\usepackage{amssymb}
\usepackage{mathtools}
\usepackage{amsthm}

\usepackage[capitalize,noabbrev]{cleveref}

\theoremstyle{plain}

\theoremstyle{definition}

\theoremstyle{remark}

\usepackage[textsize=tiny]{todonotes}

\icmltitlerunning{RegExtract: PFAS Regulatory Information Extraction with LLMs}

\begin{document}

\twocolumn[
  \icmltitle{RegExtract: A Benchmark for PFAS Regulatory Information Extraction with LLMs}

  \icmlsetsymbol{equal}{*}

  \begin{icmlauthorlist}
    \icmlauthor{Defne Circi}{yyy}
    \icmlauthor{Krystelle Lionti}{comp}
    \icmlauthor{Sarathkrishna Swaminathan}{comp}
    \icmlauthor{Daniel P Sanders}{comp}
    \icmlauthor{L. Catherine Brinson}{yyy}
    \icmlauthor{Bhuwan Dhingra}{yyy}
    \icmlauthor{Basel Shbita}{comp}
    \icmlauthor{Mohab Elkaref}{sch}
    \icmlauthor{Sara Capponi}{comp}
    \icmlauthor{Tim Erdmann}{comp}
    \icmlauthor{Brandi Ransom}{comp}
  \end{icmlauthorlist}

  \icmlaffiliation{yyy}{Duke University, USA}
  \icmlaffiliation{comp}{IBM Research, USA}
  \icmlaffiliation{sch}{IBM Research, UK}

  \icmlcorrespondingauthor{Defne Circi}{defne.circi@duke.edu}

  \icmlkeywords{Machine Learning, ICML}

  \vskip 0.3in
]

\printAffiliationsAndNotice{}  %

\begin{abstract}

Extracting structured information from regulatory documents is a critical but challenging task. Relevant facts are scattered across long, legally dense texts, and traceability of extracted facts to their source is essential for verification. With increasing awareness of the environmental and health impacts of per- and polyfluoroalkyl substances (PFAS), regulatory documents have expanded significantly, varying in scope and product relevance, making interpretation challenging for manufacturers and policymakers. We introduce \textbf{RegExtract}, a benchmark for structured information extraction from PFAS regulatory documents, comprising 488 human-annotated field instances across 20 US state documents and 46 regulations. Each
field instance includes a value and a provenance entry, yielding 976 entries in total. We define a typed, multi-field schema targeting key elements such as material definitions, permitted uses, and product exemptions, and propose provenance-aware evaluation metrics tailored to regulatory text. Through experiments with open- and closed-source LLMs across prompting strategies, document condensing methods, and schema partitioning approaches, we find that document identification fields remain challenging, and that field-specific few-shot examples improve extraction.  This work lays the groundwork for AI tools for hazardous substance identification, question answering, and compliance monitoring. The code and data are available at \href{https://github.com/IBM/PFAS-regulations}{https://github.com/IBM/PFAS-regulations}.

\end{abstract}

\section{Introduction}

Extracting structured information from long, legally dense documents
is a fundamental challenge in natural language processing. Regulatory
texts are a particularly demanding case: the extraction schema is
domain-specific, relevant information is scattered across
documents, and traceability of every extracted fact back to its source
span is critical for downstream verification and compliance
\citep{chalkidis2020legal,sie2024summarizing,wu2024framework}.
While general-purpose benchmarks such as SQuAD
\citep{rajpurkar2016squad} and Natural Questions
\citep{kwiatkowski2019natural} have driven progress in open-domain
question answering, they do not capture the structured, multi-field,
provenance-grounded extraction demanded by real regulatory workflows.
Legal benchmarks such as CUAD \citep{hendrycks2021cuad} focus on clause-span retrieval from contracts, while LegalBench \citep{guha2023legalbench} evaluates diverse legal reasoning skills through classification and generation tasks. Neither, however, addresses the problem of populating a typed, multi-field schema with provenance grounding from long regulatory documents.

In this work, we address this gap in the context of per- and
polyfluoroalkyl substances (PFAS) regulation. PFAS refers to a vast
class of synthetic fluorinated compounds, numbering in the millions
\citep{dobrzynska2025}, used for decades in products ranging from
non-stick cookware and waterproof textiles to food packaging and
cosmetics. Because these chemicals are extremely persistent and
resistant to degradation, they are now globally detected in drinking
water, soil, wildlife, and human blood \citep{ackerman2025},
underscoring their pervasive environmental presence. As evidence of accumulation in ecosystems and human populations continues to grow, their environmental persistence and potential health risks have triggered increasing regulatory scrutiny \citep{fenton2021per,yu2025global}. As a result, governments and agencies worldwide have produced a large and heterogeneous body of legislation that varies significantly in scope, language, and structure \citep{brennan2021trends}.

These regulations are embedded within long, legally dense documents,
making it difficult for manufacturers, policymakers, and compliance
officers to stay informed and compliant \citep{rademakerregulations}. Manual review of regulatory announcements is labor-intensive, error-prone, and does not scale with the increasing volume of regulatory content
\citep{demasi2022safety}. Automated extraction of structured regulation
records, including regulated products, responsible parties, enforcement
actions, and applicable exemptions, would directly support compliance
monitoring, auditing, and policy analysis.

Earlier information extraction systems relied on rule-based, statistical and deep learning based approaches \citep{korger2021rule,bommarito2021lexnlp,
zadgaonkar2021overview}. SOFIE \citep{SuchaekSOFIE} combined
pattern-based extraction with logical reasoning to extend knowledge
bases while maintaining ontological consistency. DeepDive
\citep{DeSaDeepdive} advanced this with a declarative, probabilistic
approach to knowledge base construction. Large language models (LLMs)
have since shown strong generalization to new extraction schemas via
prompting \citep{xu2023large,brown2020language}, but their reliability
on technical regulatory text remains poorly
understood \citep{kosonocky2024mining,yue2024lawllm,oliveira2024combining}.
nanoMINER \citep{OdobeskuNanoMINER} is a recent example of
LLM-based structured extraction applied to scientific documents;
our work applies a similar paradigm to the regulatory domain.
 
To address this, we introduce \textbf{RegExtract}, the first benchmark
for structured information extraction from PFAS regulatory documents,
with a focus on provenance grounding. Our key contributions are:

\begin{itemize}
    \item We present \textbf{RegExtract}, a benchmark of 488 human-annotated field instances (976 value and provenance entries) across 20 US state PFAS regulatory documents containing 46 regulations.
    \item We design a suite of prompting strategies and schema partitioning methods, evaluated across six open- and closed-source LLMs.
    \item We introduce provenance-aware evaluation metrics that jointly
    assess extraction correctness and source traceability at field,
    regulation, and document levels.
\end{itemize}
 
This work lays the foundation for trustworthy AI systems for
regulatory compliance and auditing. The schema design and evaluation
framework are general and extensible to other regulatory domains with
minor adaptation.

\section{PFAS Regulation Information Extraction Benchmark (RegExtract)}

Our approach is designed to directly resolve the three gaps motivating this work:
(1) lack of a structured, provenance-grounded benchmark $\rightarrow$ we
establish the RegExtract dataset and schema; (2) lack of a
traceable evaluation methodology $\rightarrow$ we design
provenance-aware evaluation metrics; and (3) unknown
reliability of LLMs on regulatory text $\rightarrow$ we conduct a systematic evaluation across multiple models, prompting strategies,
and document-condensing methods. We elaborate on each of these throughout this section.

\begin{figure*}[t]
  \includegraphics[width=0.95\linewidth]{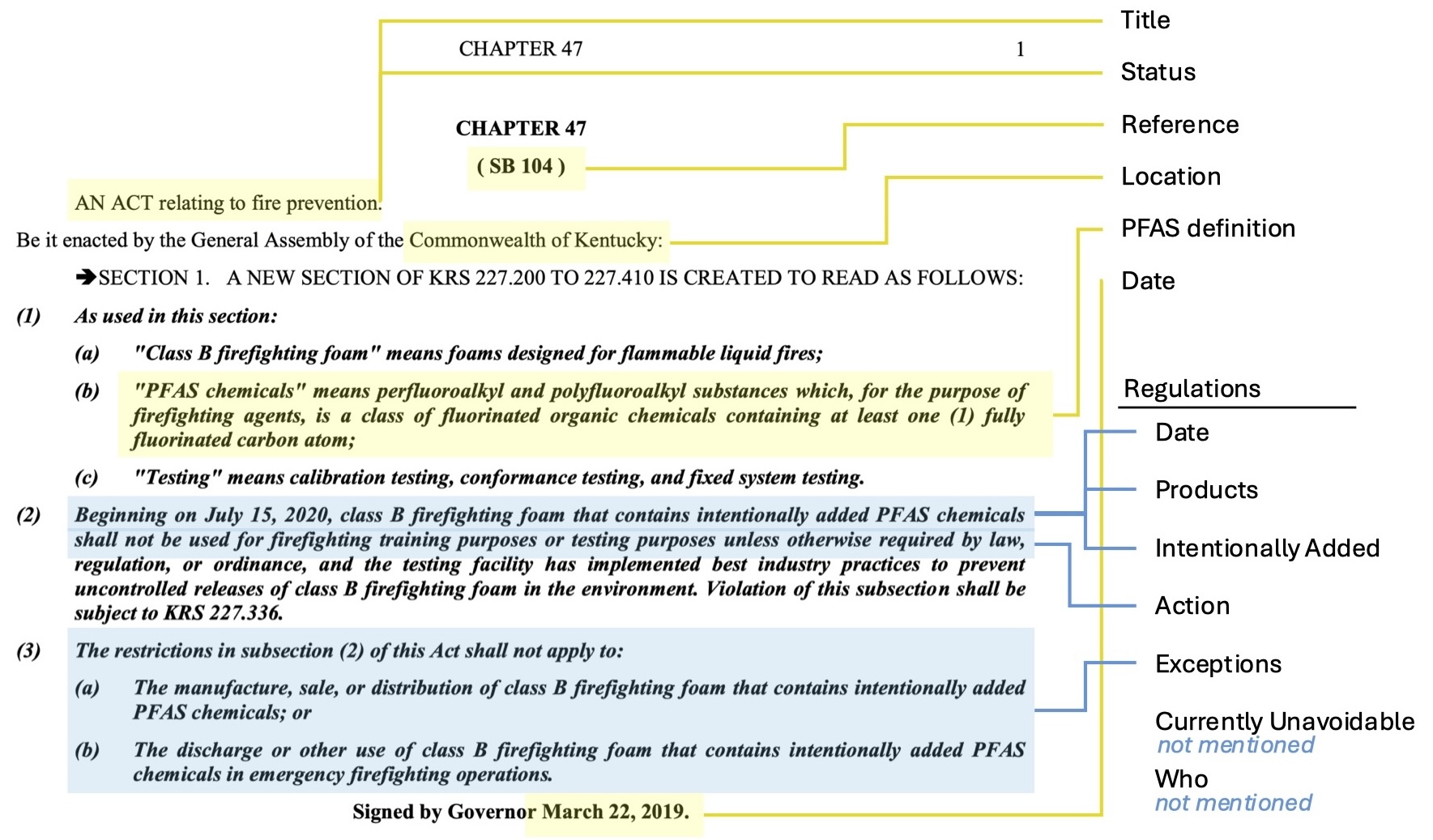}
  \caption{An example document where the highlighted text  should be extracted for each field.}
  \label{fig:schema}
\end{figure*}

We constructed a dataset of 20 US-state PFAS regulatory documents from 16 states, averaging 2.3 regulations per document, to define and extract key characteristics. The chosen fields and examples of the document text are shown in Figure \ref{fig:schema}.  Data is extracted in a structured format, with provenance indicated by the original source sentence or paragraph. Document lengths in tokens can be found in Figure \ref{fig:tokencount} in the Appendix.  The dataset was annotated by two domain experts through an iterative process: independent annotation of an initial subset of documents, comparison of results, discussion to resolve disagreements, and refinement of field definitions until consensus. The schema comprises 14 fields across document and regulation
levels: three categorical, two Boolean, four list, and five text fields.

\subsection{Problem Definition} 
We define our dataset as \( \mathcal{D} = \{D_1, D_2, \ldots, D_{20}\} \), where each \( D_i \) is a regulatory document. Corresponding to each paper \( D_i \), there is an associated list of regulations \( \mathcal{R}_i \), comprising various regulations. Formally, \( \mathcal{R}_i \) is defined as \( \mathcal{R}_i = \{r_{i1}, r_{i2}, \ldots, r_{in_i}\} \), where \( r_{ij} \) represents the \( j \)-th regulation in the regulation list of the \( i \)-th document, and \( n_i \) denotes the total number of regulations in \( \mathcal{R}_i \). Each regulation \( r_{ij} \) is represented as a structured JSON object comprising eight fields: \texttt{type}, \texttt{date}, \texttt{who}, \texttt{action}, \texttt{products}, \texttt{exceptions}, \texttt{currently\_unavoidable}, and \texttt{intentionally\_added}. Additionally, each document includes six metadata fields: \texttt{title}, \texttt{location}, \texttt{date}, \texttt{reference}, \texttt{status}, and \texttt{pfas\_definition}.

All fields are annotated with both a \texttt{value} and a corresponding \texttt{provenance}. The \texttt{value} represents the information extracted from the text, while the \texttt{provenance} provides the exact span of text in the original document that supports or contains that information. This design ensures traceability and supports downstream tasks such as information verification or model evaluation. The fields are extracted into a JSON format with value and provenance for each field, and regulations allowing for a list of objects containing the subfields under `Regulations'. The Appendix provides annotation guidelines and dataset statistics.

\subsection{Information Extraction from PFAS Regulatory Documents with LLMs} 

We leverage LLMs to extract structured regulation lists and document-level metadata from PFAS regulatory texts related to PFAS. Our methodology explores multiple dimensions of prompt engineering and information extraction strategies. Our data collection and evaluation are designed to answer the following research questions:

\begin{itemize}
\item \textbf{RQ1: Condensing the Documents} What is the best way to retrieve relevant paragraphs or chunks for classification and extraction tasks: using LLMs \cite{singh2024chunkrag} or using retrieval augmented generation (RAG)\cite{lewis2020retrieval}?

\item \textbf{RQ1.1: Chunk Selection} How should we select text chunks for processing? 
\item \textbf{RQ1.2: Document Length Impact} How does document length affect extraction and retrieval performance?
\item \textbf{RQ1.3: Embedding Choice} How do different embedding types influence the performance of retrieval-based methods? 
\item \textbf{RQ2: Extraction Strategy} What is the most effective approach for extracting values and provenance? Should values and provenance be extracted together or separately?
\item \textbf{RQ3: Schema Partitioning} Should document-level and regulation-level information be extracted together or separately?
\item \textbf{RQ4: Prompt Design} How does example usage within the prompt structure impact extraction accuracy? 

\end{itemize}

\subsection{Evaluation Metrics}
The goal of this work is to create a pipeline which can accurately extract, document, and interpret a regulation text, while ensuring the model references the source text rather than relying on its training data. We evaluate model performance at both value and provenance levels, using metrics appropriate to each field's data type. The following sections detail the evaluation criteria for each field category.

\noindent\textbf{Document Condensing Evaluation.}
The goal of condensing our documents for question RQ1 is to retain all necessary paragraphs to fill out the fields in our schemas. We calculate the rougeL recall of provenance strings contained within the condensed version of the document text. 

\noindent\textbf{Value-level Evaluation.}

\begin{itemize}
    \item Categorical fields: For the fields \texttt{type}, \texttt{pfas\_definition}, and \texttt{status}, the model is instructed to choose from a list of strings for each field. Model predictions are evaluated based on exact string match.
    \item List fields: For fields within the regulations including \texttt{who}, \texttt{action}, \texttt{products}, and \texttt{exceptions}, we calculate the $F_{1}$ score comparing the predicted list to the ground truth, treating each unique item as a binary prediction (present or absent).
    \item Boolean fields:  For \texttt{currently\_unavoidable} and \texttt{intentionally\_added}, the model is instructed to output \texttt{True} or \texttt{False}. Predictions are evaluated by checking equality with the ground-truth values.
    \item Text fields: We evaluate \texttt{title} values using ROUGE-Lsum $F_1$ with stemming \citep{lin2004rouge} to measure lexical overlap with the ground truth. Values for \texttt{location}, \texttt{reference}, and \texttt{date} are evaluated using case-insensitive exact matching.
\end{itemize}

\noindent\textbf{Provenance-level Evaluation.}
The provenance score evaluates how well the predicted provenance matches the ground truth provenance, which is a text span directly extracted from the source document. We evaluate non-missing provenance spans using ROUGE-Lsum $F_1$ with stemming \citep{lin2004rouge}, measuring lexical overlap between the predicted and ground-truth spans. We use this metric because the goal of the provenance field is faithful extraction of the supporting text, rather than paraphrasing. Semantic similarity metrics would reward paraphrases, whereas our evaluation intentionally favors high lexical overlap with the original text to ensure that the model provides verifiable evidence from the document. If the ground truth does not provide provenance, the prediction should output “not specified” in the provenance section.

\noindent\textbf{Regulation-level Aggregation.}
Documents contain one or more PFAS regulations, and the ground truth contains the relevant regulations. Value and provenance scores are averaged over matched regulations within each document, then across valid documents for each configuration. We align predicted and ground-truth regulations using greedy similarity-based matching. Processing ground-truth regulations in their existing order, we select the unmatched prediction with the highest average value-field similarity. Pairwise similarity between a predicted and ground truth regulation is computed as the mean similarity across all value fields, using the per-field metrics defined above. Once a match is established, it is removed from further consideration. We then report precision, recall, and $F_{1}$ for regulation-level matching, which measures the system's ability to \emph{identify} individual regulations. This is distinct from value and provenance scores, which measure the accuracy of the \emph{content} within each matched regulation. This aggregation method ensures a fine-grained evaluation that allows for separate analysis of regulation identification versus the correctness of extracted content and evidence grounding. For each matched pair, we compute value and provenance scores as described in the previous `Value-level Evaluation' and `Provenance-level Evaluation' sections.

\noindent\textbf{Document-level Metadata Evaluation.}
For document-level metadata fields, which are \texttt{title}, \texttt{location}, \texttt{reference}, \texttt{status},
\texttt{date}, and \texttt{pfas\_definition}, we apply evaluation metrics consistent with the field types as mentioned in the Value-level Evaluation subsection. These scores are computed per field and averaged across the dataset. When a prompting strategy produced an invalid output for a document (e.g., malformed JSON), that document was excluded from metric averages for the corresponding strategy.

\begin{figure}[t]
    \centering
    \includegraphics[width=\columnwidth]{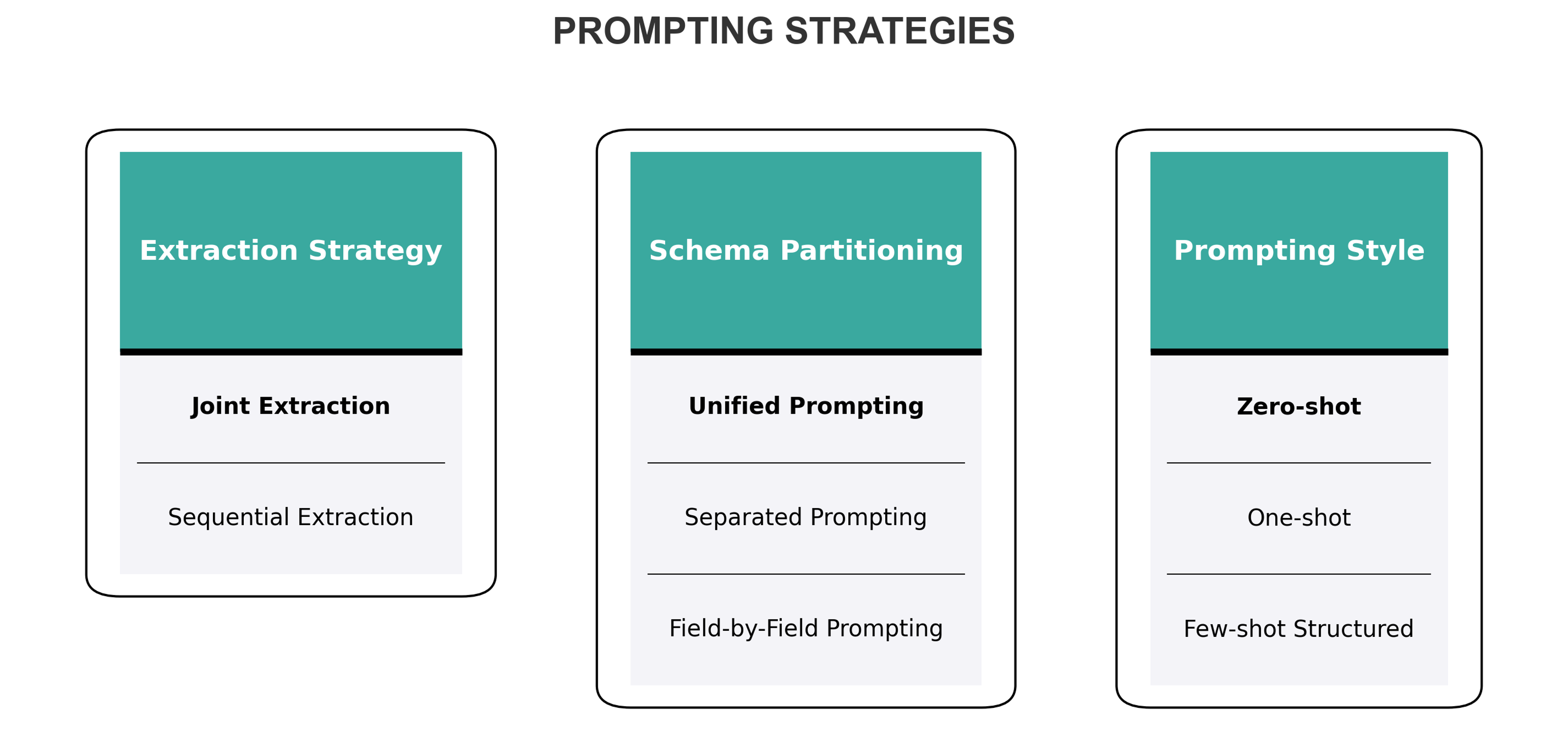}
    \caption{Prompting strategies for regulation extraction with LLMs.}
    \label{fig:prompt}
    \vspace{-13pt}
\end{figure}

\subsubsection{Design of Experiments}

We evaluate six models (GPT-4o, LLaMA 3.3 70B, LLaMA 4 Maverick, Granite-3.3-8B, Granite-4.0-H-Micro, and Mixtral-8x22B) on the full RegExtract dataset, with document condensing additionally tested across
three embedding models and multiple chunk sizes/counts. Full
hyperparameters, embedder details, and chunk configurations are listed
in Appendix~\ref{app:details}. In our first experiments for RQ1-1.3, we tested the LLM classifier and
RAG with varying numbers of chunks. Our joint extraction (baseline)
prompt is created with the version of RQ2-4 which matches the
highlighted parts in Figure \ref{fig:prompt}. We adjust the prompt to
test a single RQ, holding the sections of the prompts pertaining to
other RQs steady. For example, the Extraction Strategy was tested by
keeping the Data Separation and Prompt sections as their defaults
(highlighted), and only changing the Extraction Strategy section of
the prompt. These subtle prompt changes are outlined in the supplementary information. This design of experiments significantly
lessens the computational burden to only 6 pipelines to run on all 20
documents and allows us to isolate the interesting effects outlined in
the RQs. All prompts are a combination of one from each section in
Figure~\ref{fig:prompt}. The baseline prompt referred to as 'Joint
Extraction' includes joint extraction, unified prompting, and
zero-shot sections of the prompting strategies.

\section{Results}

\subsection{RQ1: Condensing the Documents}

We compared RAG and LLM-based document condensing across varying chunk
sizes and numbers (full results in Appendix
Figures~\ref{fig:RagRougeResults} and~\ref{fig:LLMRougeResults}).
Scores trend positively with chunk size and number for both methods.
For RAG, there is not a clear winner between embedding models; all
three perform similarly. The best configuration is 10 chunks
at 256 tokens. For LLM classifiers, 512 tokens is the best performing
chunk size, with GPT-oss-20b, llama3-3-70b-instruct, and Granite-4
as top models. Interestingly, several models get worse at 1024 tokens,
suggesting that larger chunks prevent the LLM from discarding
irrelevant information co-located with relevant content.

\begin{figure*}[h]
    \centering
    \includegraphics[scale = 0.275]{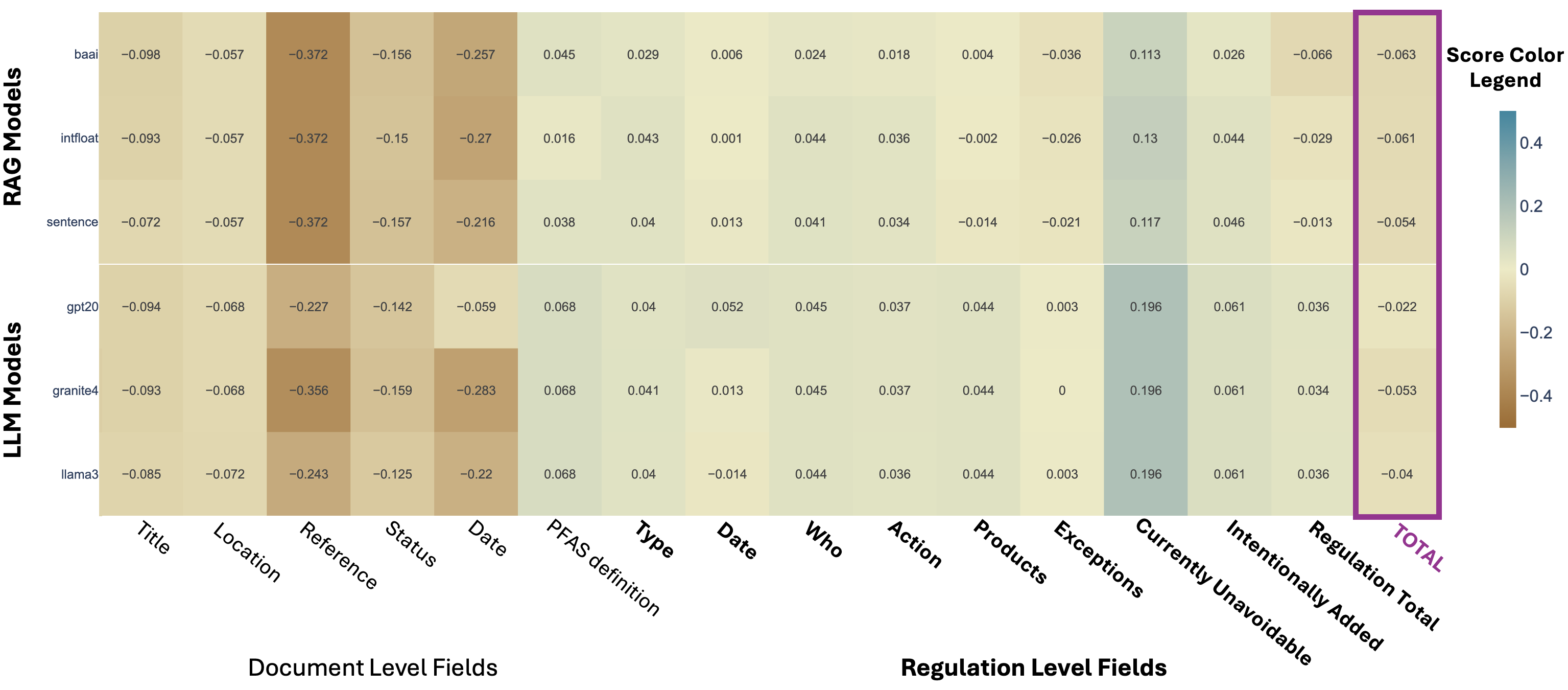}
    \caption{These heatmaps represent the recall rougeL scores calculated on the provenance of each field outlined in the schema, in comparison to the scores on the full text. Higher scores (more blue) indicate better provenance retention. The scores averaged across all documents for each text after condensed by the top performing models from the LLM protocol with chunk size of 512 tokens (top rows) and RAG protocol with chunk size of 256 tokens and 10 chunks (bottom rows).}
    \label{fig:FieldRougeResults}
\end{figure*}

The per-field scores for the best configurations are shown in
Figure~\ref{fig:FieldRougeResults}. Overall scores are within
0.03--0.06 of the full-document baseline. Document-level fields \texttt{reference}, \texttt{status}, and
\texttt{date} show larger decreases in provenance retention,
whereas \texttt{pfas\_definition},
\texttt{currently\_unavoidable}, and \texttt{intentionally\_added}
have the highest retention scores. One possible explanation is
that condensing favors passages directly related to PFAS provisions,
while document-level metadata may appear separately from these
passages and therefore be omitted. Both methods reduce the amount of text supplied to the extraction
model (Figure~\ref{fig:LengthScatter}). Such reductions may be useful for processing longer documents, consistent with prior work on query-based summarization of legal contracts
\cite{zin2023information}.

\subsection{RQ2,3,4: Information Extraction Results}
\begin{figure*}[!t]
    \centering
    \includegraphics[width=\textwidth]{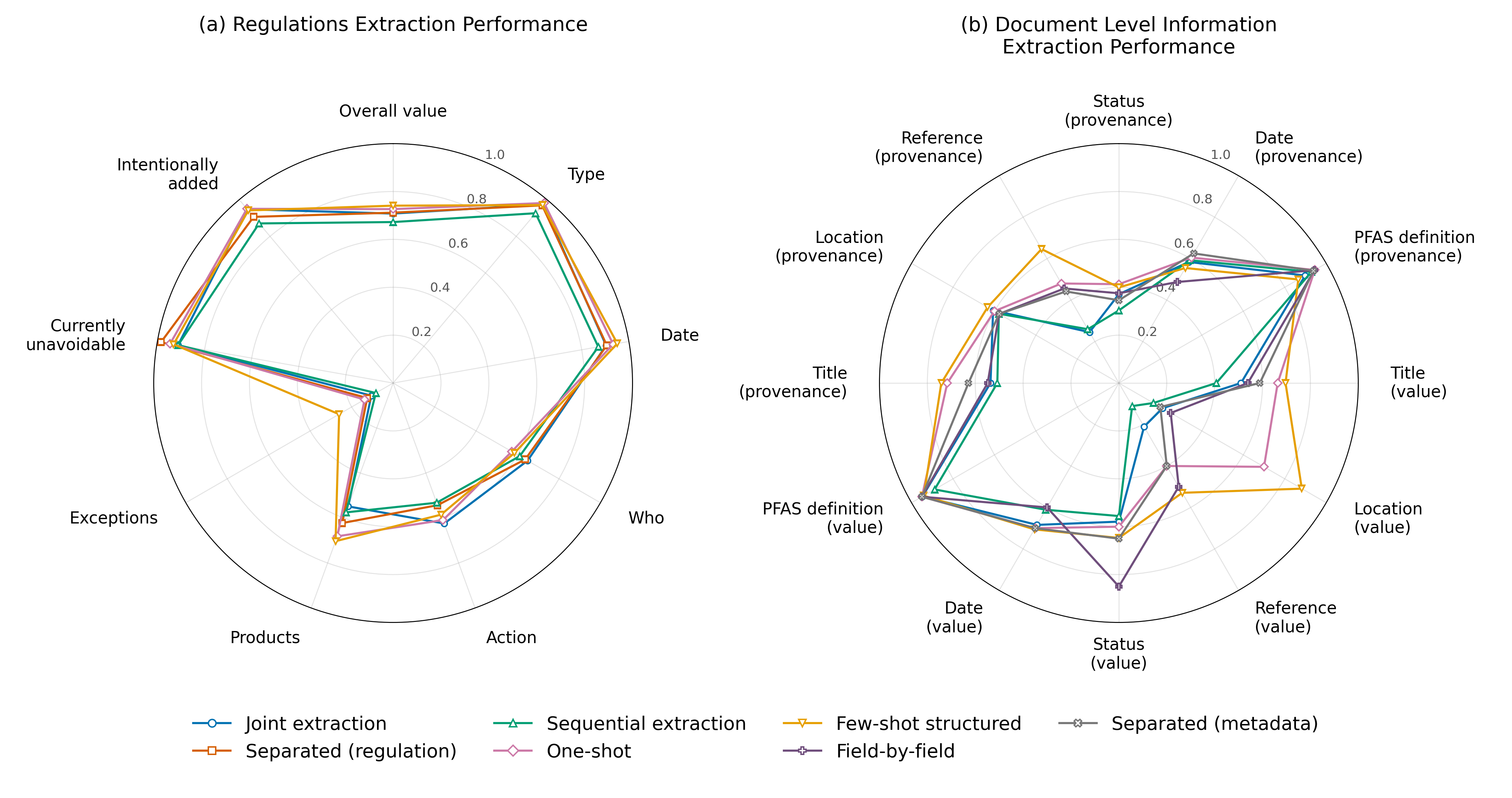}
    \caption{GPT-4o extraction performance on full documents under
    different prompting strategies. (a) Average value scores for eight
    regulation fields and their overall average. (b) Average value and
    provenance scores for six document-level fields. Scores are averaged across valid documents for each strategy; higher scores indicate better performance.}
    \label{fig:full_text_radar}
\end{figure*}

We evaluated six prompting strategies in total. The number of valid document outputs for each configuration is summarized in Table \ref{tab:pfas_prompt_comparison}. Most prompts produced valid outputs for the majority of documents, with the lowest count being 16/20 for the separated (regulation) prompting setup. Figure \ref{fig:full_text_radar} shows performance results for all fields, for all experiments carried out on full documents using GPT-4o. Table \ref{tab:pfas_prompt_comparison} shows the regulation-level aggregated results.

For the \textbf{extraction strategy}, joint extraction, which predicts values and provenance in a single step, was slightly faster than sequential extraction (27.48 versus 28.14 seconds per document). Sequential extraction achieved a higher regulation-level matching $F_{1}$ score (0.94 versus 0.88), whereas joint extraction achieved a higher aggregate regulation
value score (0.71 versus 0.67). Provenance scores were similar. Document-level performance varied by field, with neither strategy consistently outperforming the other.

Under \textbf{schema partitioning}, separated prompting was on par with joint extraction, producing stronger regulation-level $F_{1}$ scores while maintaining reasonable runtime efficiency (8.59 seconds for document-level metadata prompts and 27.24 seconds for regulation-level separation). Interestingly, the separated prompting had the same value score with lower provenance scores, either highlighting possible gaps in provenance formatting or showing that the LLM was `lucky' and found the value from the wrong provenance. Field-by-field prompting for metadata averaged 9.57 seconds and did not offer clear overall advantages, though it performed better on certain metadata categories such as \textit{status}. Through the results of the field-by-field prompting, we are able to see that most fields require further context in order to be extracted.

In terms of \textbf{prompting style}, few-shot structured prompting achieved the best overall balance in our experiments, attaining the highest average value and nearly highest provenance scores across all strategies (32.14 seconds on average). 

All experiments produce equal or higher $F_{1}$ regulation matching results than the baseline. Across all experiments, the strategy that performed the best in regulation matching, did not perform best in the values and provenance, which are of higher importance. In reviewing the scores in Table \ref{tab:pfas_prompt_comparison}, combining sequential extraction with the few-shot structured would be a promising direction.

In Tables \ref{tab:llm_metrics_basic} and \ref{tab:llm_metrics_reg_avg}, we show the results of models when using few-shot prompting. The models show variation in performance across individual fields. Under few-shot prompting, RAG-condensed inputs achieved a
regulation-level matching $F_{1}$ of 0.92, compared with 0.88 for both full-text and LLM-condensed inputs. For aggregate regulation value scores, full text performed best (0.74), followed by LLM-condensed (0.71) and RAG-condensed inputs (0.70). For the document-level fields, the results are not only scattered but show significant gaps for specific fields. Among the six models evaluated on full text, Llama 4 achieves the highest location-value score of 0.95, but ties with Granite 4 for the lowest reference-value score of 0.40.

Additional results for RAG-condensed inputs are presented in
Figures~\ref{fig:schema_regulation_radar_plot_rag}
and~\ref{fig:schema_metadata_radar_plot_rag} and
Table~\ref{tab:pfas_prompt_comparison_after_RAG}.
Corresponding results for LLM-condensed inputs appear in
Figures~\ref{fig:schema_regulation_radar_plot_llm}
and~\ref{fig:schema_metadata_radar_plot_llm} and
Table~\ref{tab:pfas_prompt_comparison_after_LLM}.
Document-level performance varies across fields. \textit{PFAS definition} scores are consistently high, while \textit{Reference} scores are lower and document \textit{Date} scores vary across prompting strategies and condensing methods. These results suggest that condensing can reduce input length while retaining useful information, although extraction performance varies across fields and configurations.

Overall, these results suggest that separating task focus and incorporating structured exemplars enhance both the completeness and reliability of PFAS regulation extraction, with moderate increases in computational cost. Model performances on individual regulation fields and evaluation of prompt strategies on the condensed documents can be found in the appendix.

\begin{table*}[t]
\caption{Summary of evaluated prompts with gpt-4o, total valid document outputs, and average processing time (s) per configuration. Metrics include regulation-level Precision (P), Recall (R), and $F_{1}$, along with averaged value and provenance scores for regulation fields. Scores are averaged across valid documents. Field-by-field  and Separated (Document-level) are not applied within regulations.}
\centering
\small
\begin{tabular}{lcc|ccccc}
\hline
\textbf{Prompt Strategy} & \textbf{\#Valid} & \textbf{t (sec)} & \textbf{Reg. P} & \textbf{Reg. R} & \textbf{Reg. }\textbf{$F_{1}$} & \textbf{Avg. Val. Score} & \textbf{Avg. Prov. Score} \\
\hline
Joint Extraction (baseline)   & 19 & 27.48   & \textbf{1.00} & 0.83 & 0.88 & 0.71 & \textbf{0.65} \\
Sequential Extraction  & 18 & 28.14  & \textbf{1.00} & 0.91 &\textbf{ 0.94} & 0.67 & 0.64 \\
Separated (Regulation)  & 16 & 27.24  & 0.94 & \textbf{0.96} & 0.93 & 0.71 & 0.61 \\
One-Shot              & 20 & 39.66   & 0.98 & 0.89 & 0.91 & 0.73 & 0.59 \\
Few-Shot Structured    & 17 & 32.14  & 0.97 & 0.85 & 0.88 & \textbf{0.74} & 0.64 \\
Separated (Document-level) & 20 & 8.59 & & & & & \\
Field-by-Field & 20 & 9.57 & & & & & \\
\hline
\end{tabular}
\label{tab:pfas_prompt_comparison}
\end{table*}

\begin{table*}[t]
\caption{Performance metrics using few-shot structured prompt for various LLMs. Each document-level field shows the value and provenance score. }
\centering
\small
\begin{tabular}{l c c c c c c c}
\hline
\textbf{Model} & \textbf{\#Valid} & \textbf{Title} & \textbf{Loc.} & \textbf{Ref.} & \textbf{Status} & \textbf{Date} & \textbf{PFAS Def.} \\
\hline
llama-3-3-70b-instruct & 17 & 0.63 / 0.68 & 0.94 / 0.63 & 0.53 / 0.64 & 0.65 / 0.36 & 0.65 / \textbf{0.71} & 0.88 / 0.85 \\
llama-4-maverick& 20 & 0.70 / 0.67 & \textbf{0.95} / 0.68 & 0.40 / 0.62 & 0.60 / 0.19 & 0.60 / 0.53 & 0.95 / 0.94 \\
gpt-4o & 17 & 0.70 / \textbf{0.74} & 0.88 / 0.63 & 0.53 / \textbf{0.65} & 0.65 / 0.40 & \textbf{0.71} / 0.55 & 0.94 / 0.87 \\
granite-3.3-8b-instruct & 20 & \textbf{0.74} / 0.53 & 0.60 / 0.53 & \textbf{0.60} / 0.31 & 0.40 / 0.17 & 0.55 / 0.50 & 0.30 / 0.77 \\
mixtral-8x22B-instruct-v0.1 & 16 & 0.51 / 0.54 & 0.88 / \textbf{0.74} & 0.44 / 0.59 &\textbf{ 0.75} / \textbf{0.43 }& 0.56 / 0.56 & 0.94 / 0.85 \\
granite-4.0-h-micro & 20 & 0.62 / 0.24 & 0.75 / 0.27 & 0.40 / 0.59 & 0.40 / 0.20& 0.55 / 0.39 & 0.75 / 0.57 \\
gpt-4o (LLM condensed) & 19 & 0.67 / 0.67 & 0.84 / 0.61 & 0.42 / 0.48 & 0.58 / 0.35 & 0.63 / 0.55 & \textbf{1.00} / \textbf{0.97 }\\
gpt-4o (RAG condensed) & 19 & 0.68 / 0.69 & 0.79 / 0.62 & 0.26 / 0.37 & 0.74 / \textbf{0.43} & 0.47 / 0.41 & \textbf{1.00} / \textbf{0.97} \\
\hline
\end{tabular}
\label{tab:llm_metrics_basic}
\end{table*}

\begin{table*}[t]
\caption{Performance of different large language models on regulation-level extraction tasks using the few-shot structured prompting approach. The table reports regulation matching precision, recall, and $F_{1}$ scores, along with value and provenance evaluation metrics. Scores are averaged across valid documents. The best-performing results for each metric are highlighted in bold.}
\centering
\small
\begin{tabular}{l c c c c c c c c c c c c c c c}
\hline
\textbf{Model} & \textbf{\#Valid} & \textbf{Reg. ($P$,$R$,$F_1$)} & \textbf{Reg. Value / Prov.} \\
\hline
llama-3-3-70b-instruct &17 & \textbf{1.00} / 0.82 / 0.88 & 0.70 / 0.55 \\
llama-4-maverick-17b-128e-instruct-fp8 & 20 & \textbf{1.00} / 0.86 / 0.91 & 0.68 / 0.57  \\
gpt-4o & 17 & 0.97 / 0.85 / 0.88 & \textbf{0.74} / \textbf{0.64 } \\
granite-3.3-8b-instruct & 20 & 0.89 / 0.83 / 0.81 & 0.64 / 0.30  \\
mixtral-8x22B-instruct-v0.1 & 16 & 0.97 / 0.86 / 0.89 & 0.69 / 0.54 \\
granite-4.0-h-micro & 20 & \textbf{1.00} / 0.71 / 0.78 & 0.62 / 0.16 \\
gpt-4o (LLM condensed) & 19 & \textbf{1.00} / 0.83 / 0.88 & 0.71 / 0.58 \\
gpt-4o (RAG condensed) & 19 & \textbf{1.00} / \textbf{0.88} / \textbf{0.92} & 0.70 / 0.60 \\
\hline
\end{tabular}
\label{tab:llm_metrics_reg_avg}
\end{table*}

\subsection{Analysis of Errors}
\label{sec:errors}
In designing the prompts for information extraction, we decided to use the same prompt for both RAG and LLM. The goal is to extract any relevant information from the document text, therefore the prompt cast a wide net. This limited the specificity of the prompt, and it has been noted previously that RAG performs better with specific prompts \citep{chan2024rq}. The counter examples that are included in the prompt may benefit LLM identification, but undermine RAG retrieval. Individualized prompts to retrieve chunks for each field may work for document level fields, but for our work would not be able to keep individual regulation connections for regulation fields. 

We identify subjectivity in the annotation of field values as one source of score differences. For example, `anti-fogging sprays and wipes' was separated into `anti-fogging sprays' and `anti-fogging wipes'. The LLM may not separate this as such and for reasons similar, we expect the reported results to underestimate the LLMs' performance of the task. The fields themselves were chosen as what would be helpful for further human processing, whether or not that information is relevant to the regulation, as we see with some fields at the document level. The ``reference" is chosen to represent a way to identify the regulatory document, and identifies the source document rather than describing a regulatory requirement. For standardization, the first reference was chosen from each document, however, there are cases where there are seemingly multiple references such as "Substitute Senate Bill No. 292" and "Public Act No. 24-59". There are no instructions for the LLM to choose one reference over the other, and it would not substantially make a difference to further human processing. This is an example of a task which could be learned, given that humans may be biased towards one reference because of their previous interactions with the documents in a way that the LLMs have not been exposed to. Beyond differences in annotation, regulatory provisions often share similar language and structure, including passages unrelated to PFAS or describing implementation details rather than distinct regulations. This similarity may make it difficult for models to distinguish relevant regulations from other provisions. Another challenge is identifying the document date. It should represent when the document was formed or the meeting about the document happened, and the date is more isolated in the document structure, usually without any label. The LLMs seem to miss the context that the first date like string is the document date, and instead find the first date which comes with context such as `Approved on:', regardless of what that approval is referring to. These examples suggest that interpreting document structure and
distinguishing the roles of similar expressions remain challenges for regulatory information extraction.

We found that some of our reference scores regarding RQ1 to be slightly lower than one, and allows for some of the relative scores to be greater than zero. The benchmark dataset JSONs contain reformatted provenance text which is condensed across lines, paragraphs and pages, which may be different from the original text. Another cause for an increase in scores after condensing could be the inclusion of header information from the chunks.

Strict matching can penalize semantically equivalent answers, so the reported scores may not fully reflect the usefulness
of extracted information. The extraction protocol for information for the subjective discrepancies outlined in this section was standardized for all documents. Therefore, we believe that regulatory language patterns could be a strong candidate for a fine-tuning task, similar to information extraction from materials science articles \cite{dagdelen2024structured}.

\section{Conclusion}
Our work explores the various processes around regulatory document information extraction, the effects of document condensing, and prompt engineering for important regulation fields. We introduce the RegExtract benchmark, comprising 976
annotations across 20 documents containing 46 regulations, to support
LLM evaluation in legal, AI, and materials research. The results show variation in extraction performance across fields,
models, and prompting strategies, with substantial room for
improvement in the accuracy of extracted values and supporting
provenance. Document condensing offers a way to reduce input length,
although its effects on extraction performance vary across fields
and configurations. Its practical benefits should therefore be
weighed against the additional processing cost and document length.
No single model or prompting strategy performs best across all
fields. These findings can inform the selection and combination of
models and prompts for regulatory information extraction pipelines.

\section{Limitations}
Our annotated dataset covers 20 regulatory documents; while annotation
density is high, some conclusions may not generalize to larger or more
diverse corpora. It only contains English data, and the data came only from the United States. In investigating using documents from other countries, we found that regulations are distributed on the national level, which provides only one document in another language, or one document in many languages. In the former case, our team does not have the linguistic skill set to professionally translate many other languages, and in the latter we would only be testing the translation capabilities of the model instead of extraction and retention. The documents were selected from saferstates.org filtered by PFAS regulations, included regulations on reporting, labeling, and banning chemicals in products, did not include documents regulating contamination limits, or regulation development timelines, documents containing amendments, and were related to the supply chain of manufacturing or selling PFAS-containing products, not documents which regulated the purchaser of such products. We hope this work encourages broader community participation in developing labeled datasets to advance research in regulatory information extraction. We believe this approach is extensible to other documents with small updates in desired category extractions and prompts, which we adjusted to the specific language used in state US documents. 

List-field evaluation uses exact item matching, so equivalent
expressions such as ``rug'' and ``rugs'' are scored as different
items. This can underestimate extraction quality for
\texttt{who}, \texttt{action}, \texttt{products}, and
\texttt{exceptions}. Future work could explore normalization or semantic matching of list items.

\section*{Impact Statement}

This work supports compliance officers, manufacturers, and policymakers in navigating regulatory documents at scale, broadening access to regulatory knowledge and supporting the identification of hazardous substances. Applying LLMs to high-stakes regulatory text carries risks of misextraction; we mitigate this by pairing every extracted value with a verifiable provenance span, and emphasize that such tools should augment, not replace, legal and compliance professionals. We do not foresee additional societal consequences specific to this work that are not already established when advancing structured information extraction and legal NLP.

\newpage

\bibliography{example_paper}
\bibliographystyle{icml2026}

\newpage

\clearpage

\appendix

\section{Dataset}
The documents were selected from saferstates.org filtered by PFAS regulations, which limits our dataset to the United States. We could not find sufficient regulations from other geographic locations that met the remainder of the requirements described in this section, which are necessary to run a comparative evaluation. The documents selected  included regulations on reporting, labeling, and banning chemicals in products, and did not include documents regulating contamination limits, or regulation development timelines, documents containing amendments. All documents were related to the supply chain of manufacturing or selling PFAS-containing products, not documents which regulated the purchaser of such products. 

\begin{figure}[H]
    \centering
    \includegraphics[scale = 0.04]{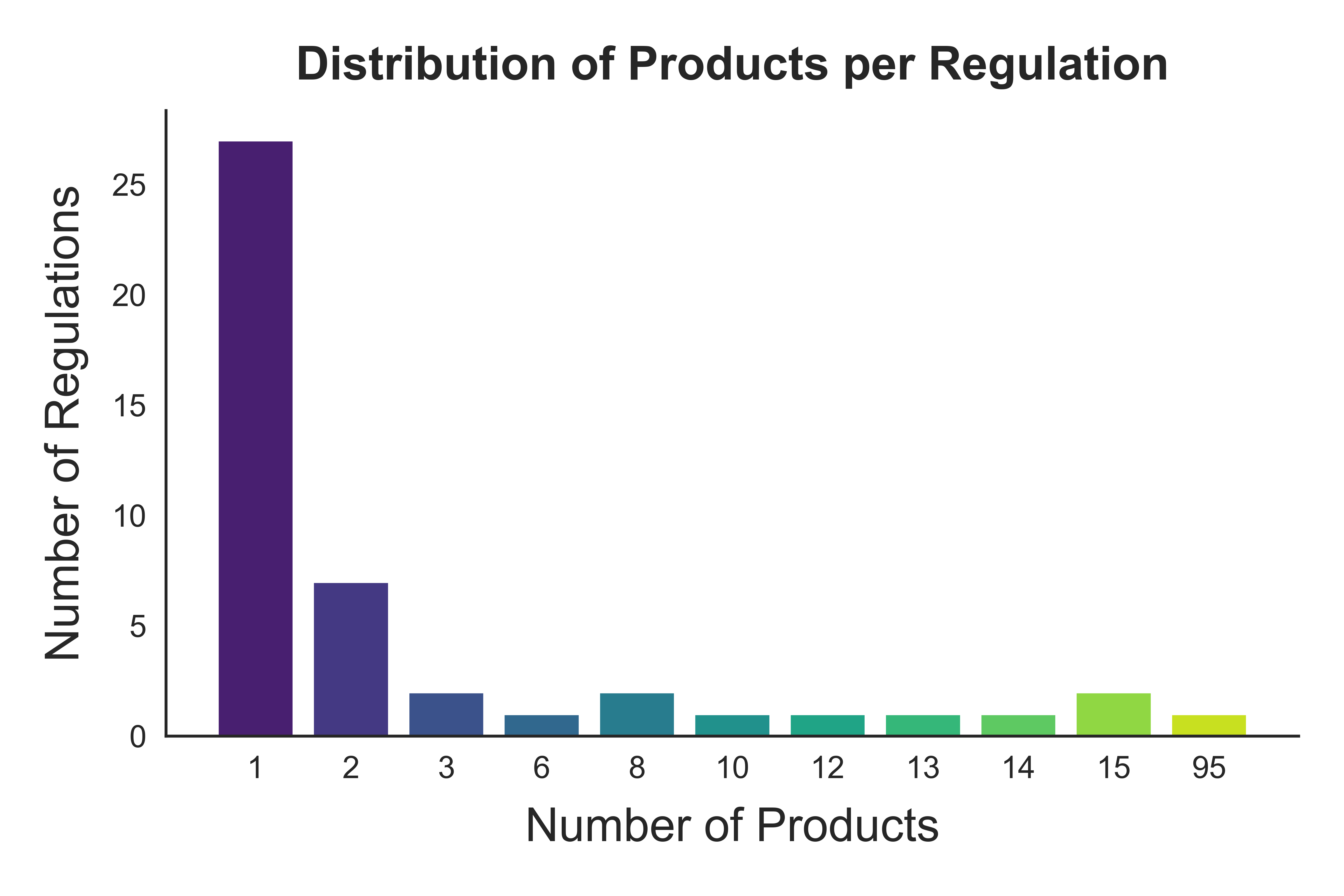}
    \caption{Products per regulation}
\label{fig:products_per_regulation_histogramlowercase}
\end{figure}

\begin{figure}[H]
    \centering
    \includegraphics[scale = 0.04]{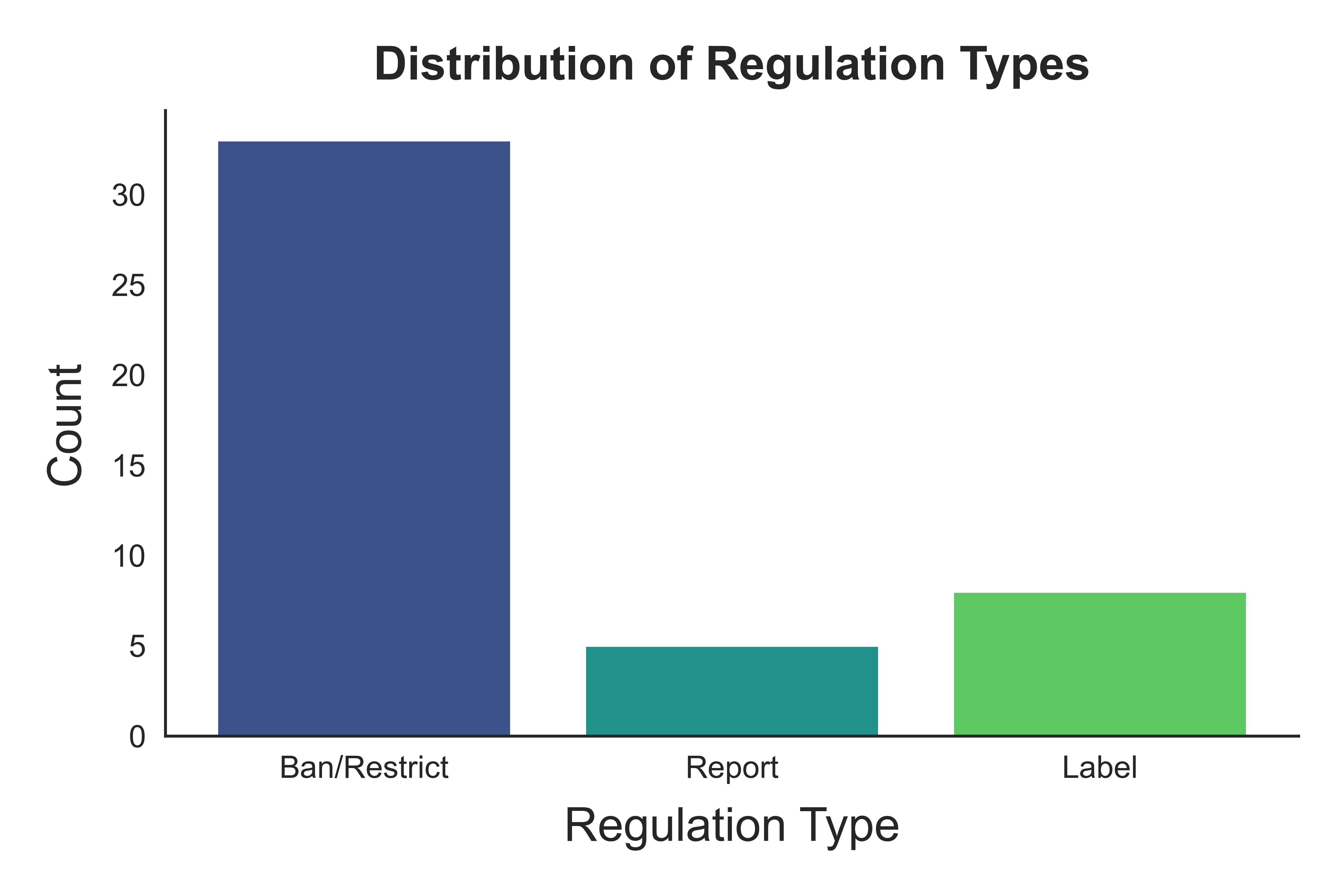}
    \caption{Regulation type distribution}
\label{fig:regulation_type_distributionKLlowercase}
\end{figure}

\begin{figure}[H]
    \centering
    \includegraphics[scale = 0.04]{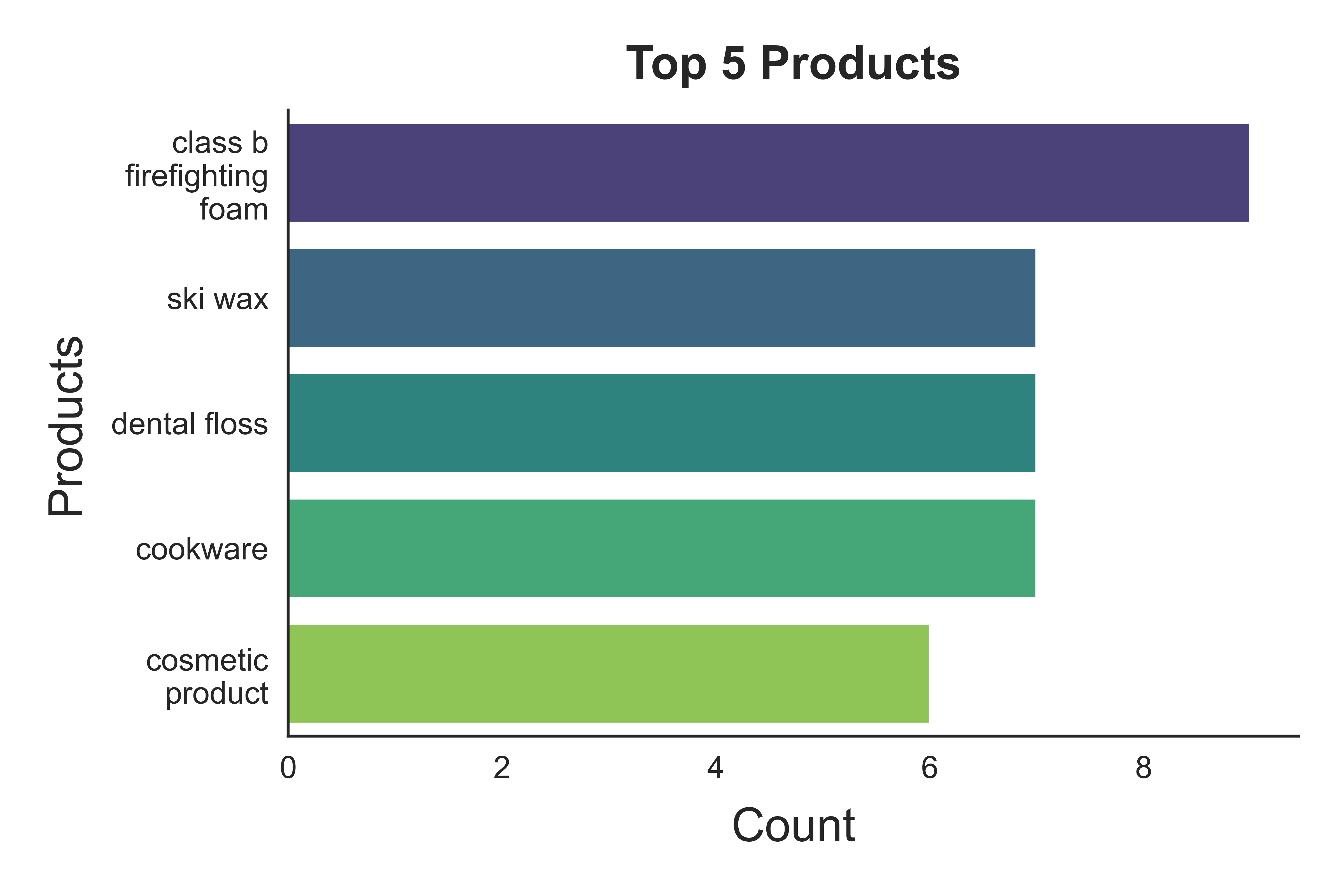}
    \caption{Top 5 products}
\label{fig:top_5_products_finallowercase}
\end{figure}

\begin{figure}[H]
    \centering
    \includegraphics[scale = 0.04]{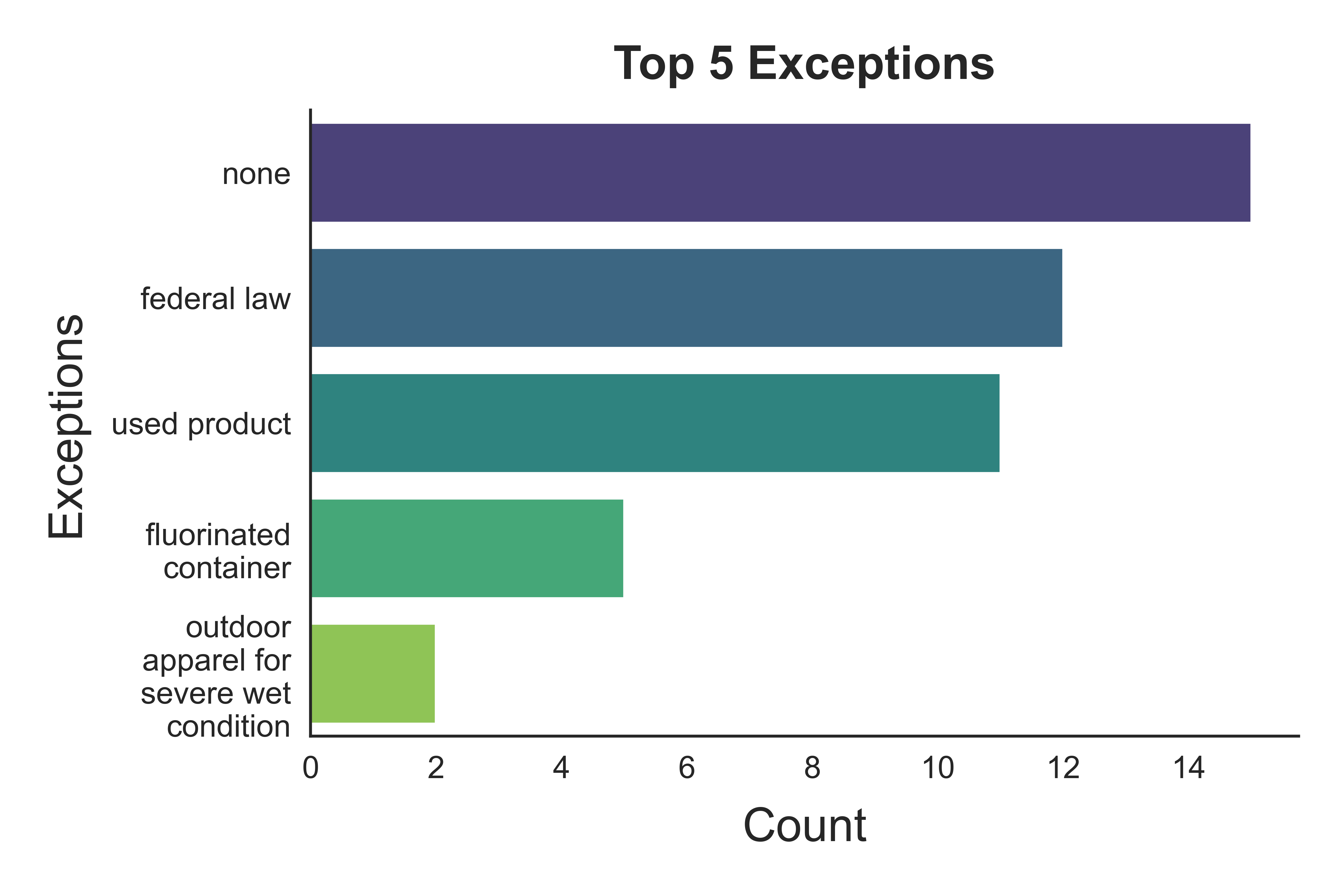}
    \caption{Top 5 exceptions}
\label{fig:top_5_exceptions_finallowercase2}
\end{figure}

\begin{figure}[H]
    \centering
    \includegraphics[scale = 0.04]{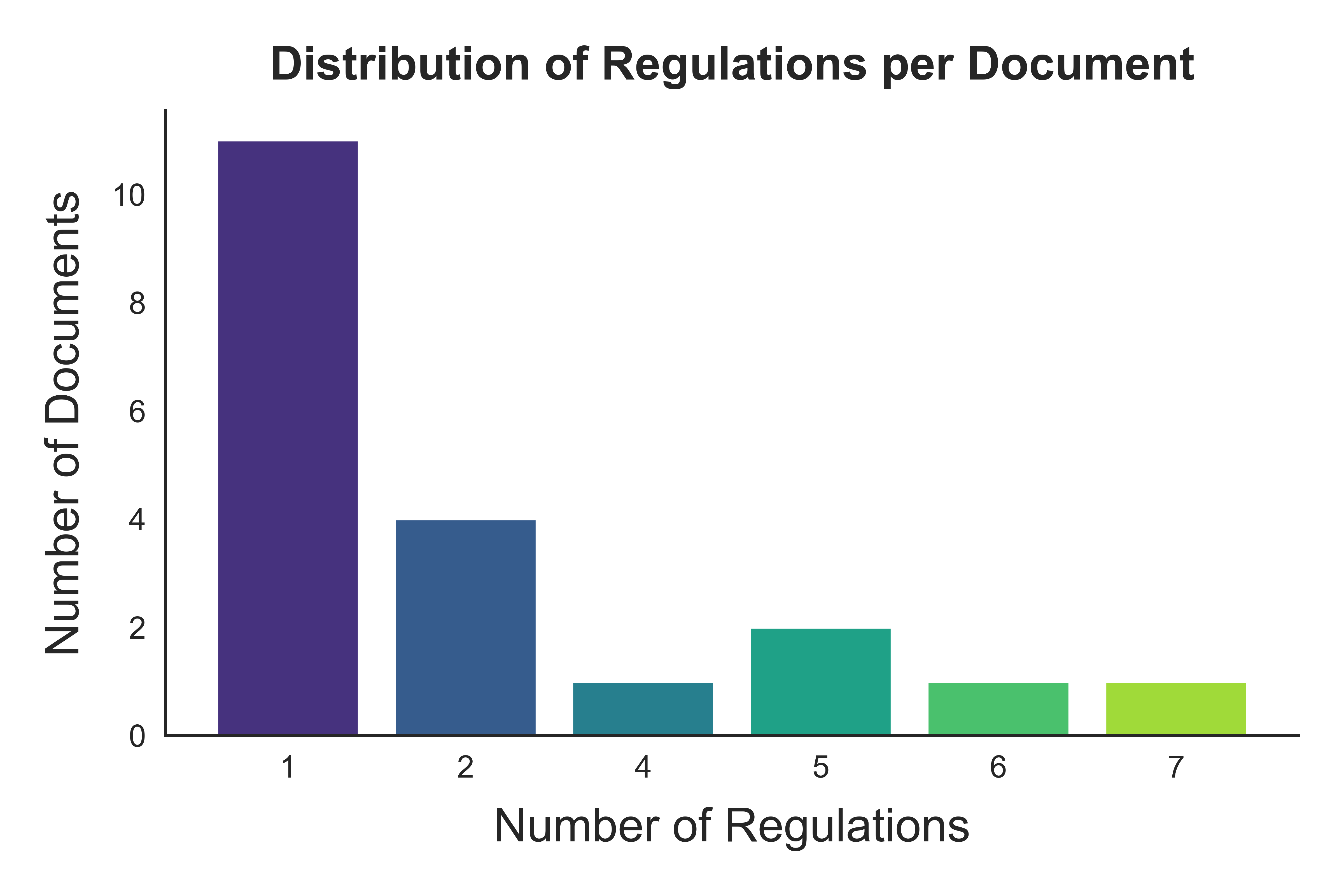}
    \caption{Regulations per file}
\label{fig:regulations_per_file_histogram_final2lowercase}
\end{figure}

\begin{figure}[H]
    \centering
    \includegraphics[scale=0.17]{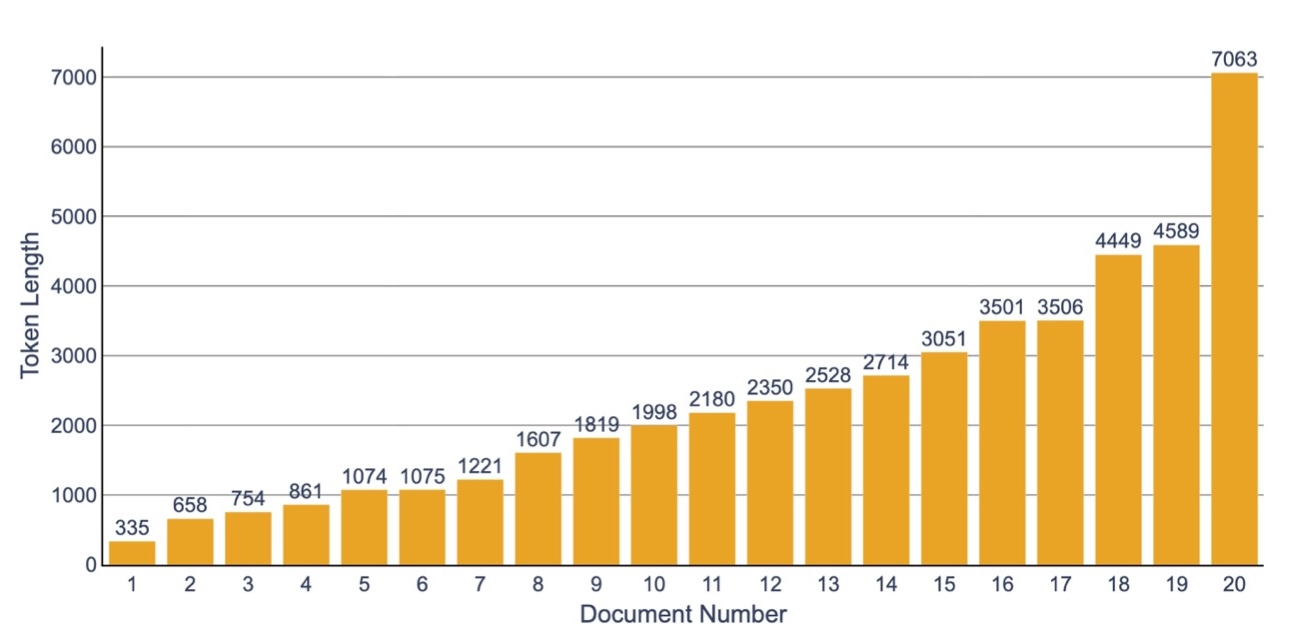}
    \caption{The length of each document using
    \texttt{AutoTokenizer.\allowbreak from\_pretrained("bert-base-uncased")}.}
    \label{fig:tokencount}
\end{figure}

The final schema for document and regulation level properties was chosen iteratively as the dataset was formed between 2 scientists. We chose to create a predefined schema in order to streamline evaluation, which we recognize limited the dataset size. Allowing a freeform schema determined by the LLM would not allow us to evaluate the retention of provenance in the document, which is a main focus of this work. The fields were chosen as the minimum number of relevant fields to define a regulation, which could be applied to all regulation types of report, label, and ban. A few of the fields require the choice from a set list of strings and some are freeform, which also allowed us to test the models' ability to accomplish different types of tasks. As far as we are aware, this is the most comprehensive dataset of annotated documents to analyze chemical regulation data extraction. Figures \ref{fig:products_per_regulation_histogramlowercase}, \ref{fig:regulation_type_distributionKLlowercase}, \ref{fig:top_5_products_finallowercase}, \ref{fig:top_5_exceptions_finallowercase2}, and \ref{fig:regulations_per_file_histogram_final2lowercase} give an overview of the types of information found in these documents and the specific fields. Figure~\ref{fig:tokencount} shows the length of each document
in tokens.


\begin{figure*}[H]
    \centering
    \includegraphics{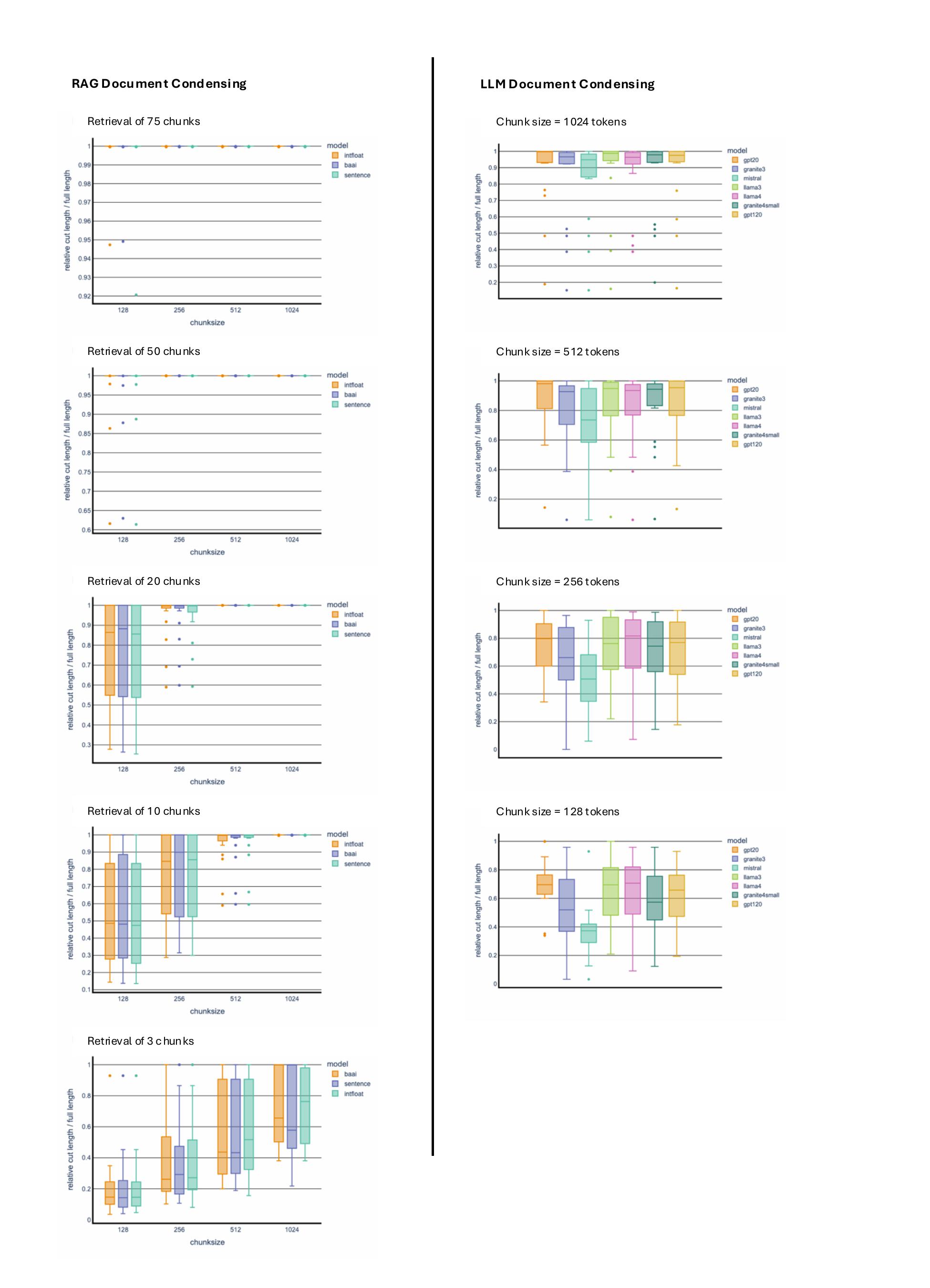}
    \caption{These data points represent the difference in the text length after the text is condensed by a RAG or LLM protocol. Each box plot represents all of the documents for a single model, using various chunk size and numbers for each of the condensing protocols. The left column shows all of the RAG variations which were tested, the right column shows all of the LLM variations which are tested.}
    \label{fig:Document_cut_lengths_overview}
\end{figure*}

\section{Details of Experiments}
\label{app:details}

For each question, we investigate a variation in prompt design and empirically evaluate the impact of these strategies:
\noindent\textbf{1. Document Processing.}
Regulatory documents can be over 100 pages long, far exceeding the context window for many models and providing mostly irrelevant information to PFAS-specific regulations. The documents are all chunked with the docling chunker \cite{livathinos2025docling}, and each chunk is put through the evaluative processes.
We investigate how we can narrow in on relevant content from the text by comparing 2 strategies for condensing the document:
\begin{itemize}
    \item LLM Classifier: We prompt an LLM on whether it thinks the chunk of text is relevant, using prompt P1 in the Supplementary Information, and concatenate selected paragraphs into a smaller document for further processing. 
    \item RAG Extraction: We embed each chunk using an embedder and return the top N chunks in terms of similarity to embedded prompt P1.
    \begin{itemize}
        \item For a fixed number of total chunks, we will also extract chunks based on context windows to assess RQ1.1
        \item From our array of document lengths in our experimental set we will assess RQ1.2
        \item We compare multiple embedders to see their effect of extraction performance to assess RQ1.3
    \end{itemize}
\end{itemize}

\noindent\textbf{2. Extraction Strategy.}
We investigate two strategies for extracting values and their corresponding provenance. This will tell us if a more efficient pipeline (fewer LLM calls) is able to keep up with the same performance. 
\begin{itemize}
\item Joint Extraction: Prompting the model to generate both the value and the provenance for each field in a single step. 
\item Sequential Extraction: First prompting for field values, then prompting separately for their provenance spans based on the predicted values.
\end{itemize}

\noindent\textbf{3. Schema Partitioning.}
The length and detail of a prompt can affect the output; therefore, we explore how the grouping of information affects extraction quality. 
\begin{itemize}
\item Unified Prompting: Extracting all document-level (metadata) and regulation-level fields together. 
\item Separated Prompting: Extracting metadata and regulation fields in independent steps and merging the outputs. 
\item Field-by-Field Prompting: Issuing separate prompts for each document-level metadata field to improve accuracy.

\end{itemize}

\noindent\textbf{4. Prompting Style.}
We experiment with prompting with different levels of examples given to the LLM in the prompt \cite{brown2020language} to evaluate the trade-off of providing more information and the LLM bias towards parroting the exact answer. 
\begin{itemize}
\item Zero-shot: Direct prompts without examples. 
\item One-shot: Prompts with a single schema example for guidance. 
\item Few-shot Structured: Prompts with a schema example as well as examples of values for each individual field.
\end{itemize}

\subsection{Design of Experiments}
For condensing the articles, we used a range of LLMs for extraction, including
LlaMA 3.3 70b, LLaMA 4 (Maverick), granite-3.3-8b-instruct, granite-4-h-small, GPT-oss-20b, GPT-oss-120b, and mixtral-8x22B-instruct-v0.1. For RAG, we use BAAI/bge-m3, intfloat/multilingual-e5-small, and sentence-transformers/all-MiniLM-L6-v2 to encode raw texts and queries. For information extraction, we report results for six models: LLaMA 3.3 70B instruct \cite{grattafiori2024llama}, LLaMA 4 (Maverick), granite-3.3-8b-instruct, granite-4.0-h-micro, GPT-4o \citep{hurst2024gpt}, and mixtral-8x22B-instruct-v0.1. We picked the max token number = 8192, and temperature = 0. 

Our first experiments for RQ1-1.3 are carried out on all 20 documents. We tested the LLM classifier and RAG with varying numbers of chunks. Our prompt P1 is created with the version of RQ2-4 which matches the highlighted boxes in Figure \ref{fig:prompt}. We adjust the prompt to test a single RQ, holding the sections of the prompts pertaining to other RQs steady. For example, the Extraction Strategy was tested by keeping the Data Separation and Prompt sections as their defaults (highlighted), and only changing the Extraction Strategy section of the prompt. These subtle prompt changes are outlined in the Supplementary information. This design of experiments significantly lessens the computational burden to only 6 pipelines to run on all 20 documents and allows us to isolate the interesting effects outlined in the RQs. 

\section{Discussion of condensing the documents}
The motivation for condensing the documents is to lessen the context that would be passed to the extraction pipeline. Regulatory documents can be up to hundreds of pages long, with multiple topics and regulations being described, and passing the full document to the extraction pipeline would far exceed the context window. The condensing methods of RAG and LLM that we use focus on the ability to isolate one topic, PFAS, and are measured by the amount of relevant information that is retained in the trimmed version of the document. The 20 documents analyzed in this article fit within the context window of all models in their entirety, so that we can compare the effects of trimming the document against the full-document baseline. 

\begin{figure*}
    \centering
    \includegraphics[scale=0.9]{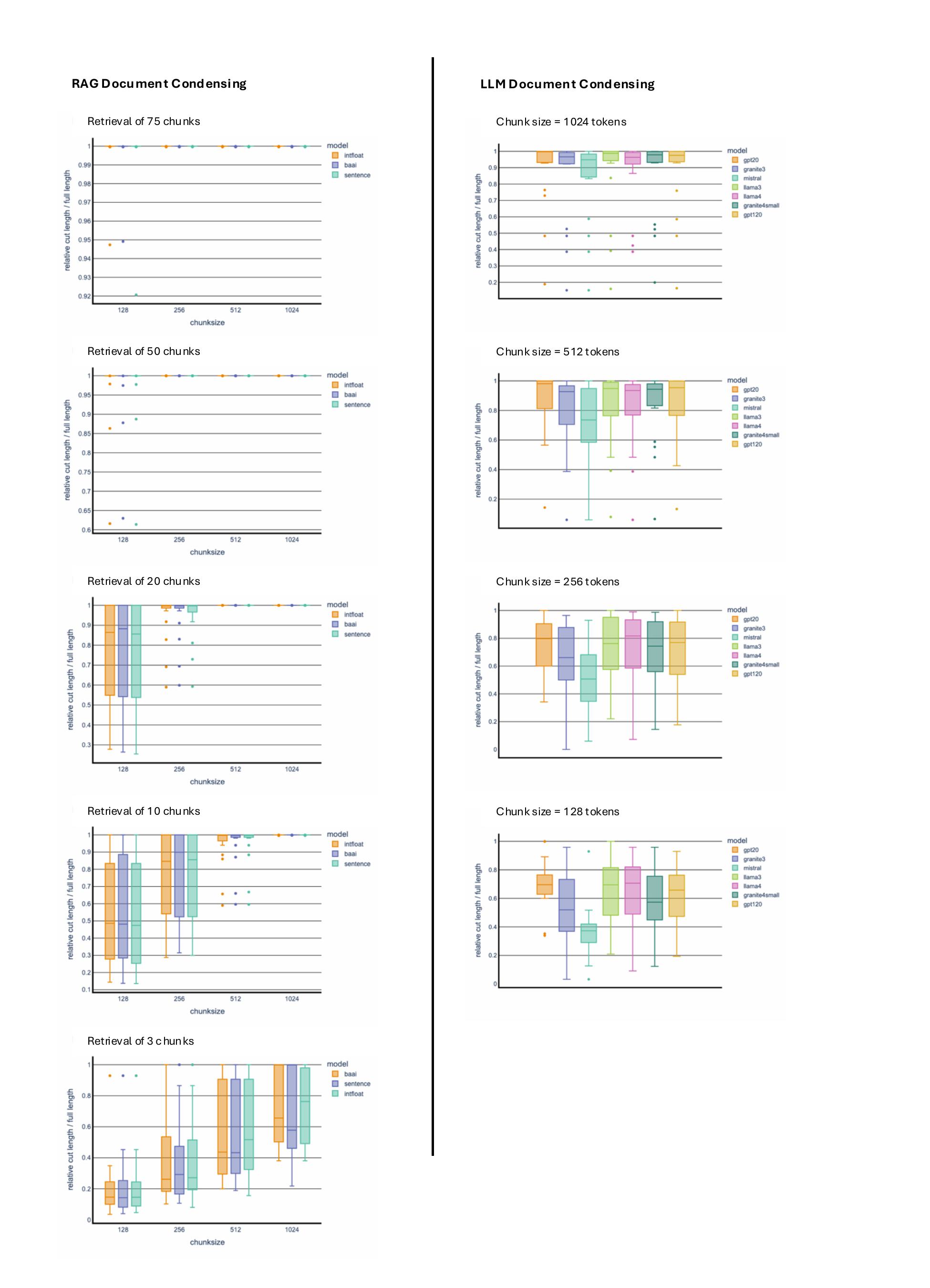}
    \caption{These data points represent the difference in the text length after the text is condensed by a RAG or LLM protocol. Each box plot represents all of the documents for a single model, using various chunk size and numbers for each of the condensing protocols. The left column shows all of the RAG variations which were tested, the right column shows all of the LLM variations which are tested.}
    \label{fig:Document_cut_lengths}
\end{figure*}

Each document is trimmed by selecting relevant chunks, produced by docling's HybridChunker, which are composed of the text and heading for that chunk. An LLM Classifier can select as many relevant chunks as it sees fit, and the RAG Extraction is limited to 3, 10, 20, 50, or 75 chunks. The effect of number and size of chunks on how much of the document is kept, measured by relative length to the full document, is shown in figure \ref{fig:Document_cut_lengths}. Here we see that, for RAG, for larger chunk sizes, it doesn't take more than 10 chunks to reach the full length of the original text for many documents. Contrasted with the LLMs, which discard parts of the document no matter what, even though they are not forced to choose a certain number of chunks.  The short length of the documents requires a fairly significant limitation in order to require the RAG process to discard data. Even with a limit of 3 chunks of 256 tokens, one of the documents is already at its full length.

\begin{figure*}
    \centering
    \includegraphics[scale=0.16]{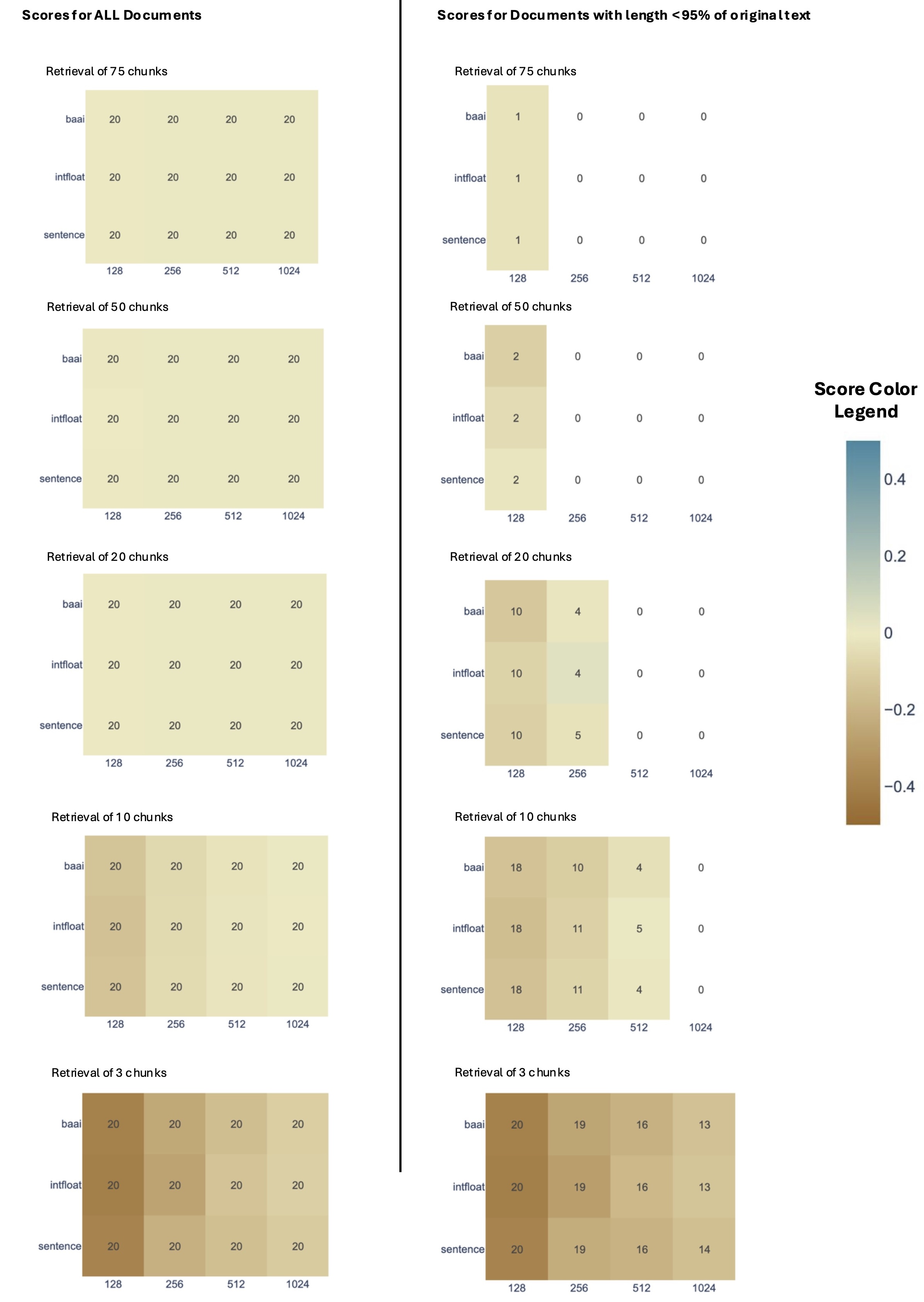}
    \caption{rovenance ROUGE-L recall scores relative to the full-text baseline after RAG condensing. Scores are averaged
    across fields and documents included in each panel. The left panel includes all 20 documents; the right includes only
    documents condensed to less than 95\% of their original length. Cell labels indicate document counts.}
    \label{fig:RagRougeResults}
\end{figure*}

\begin{figure*}
    \centering
    \includegraphics[scale=0.2]{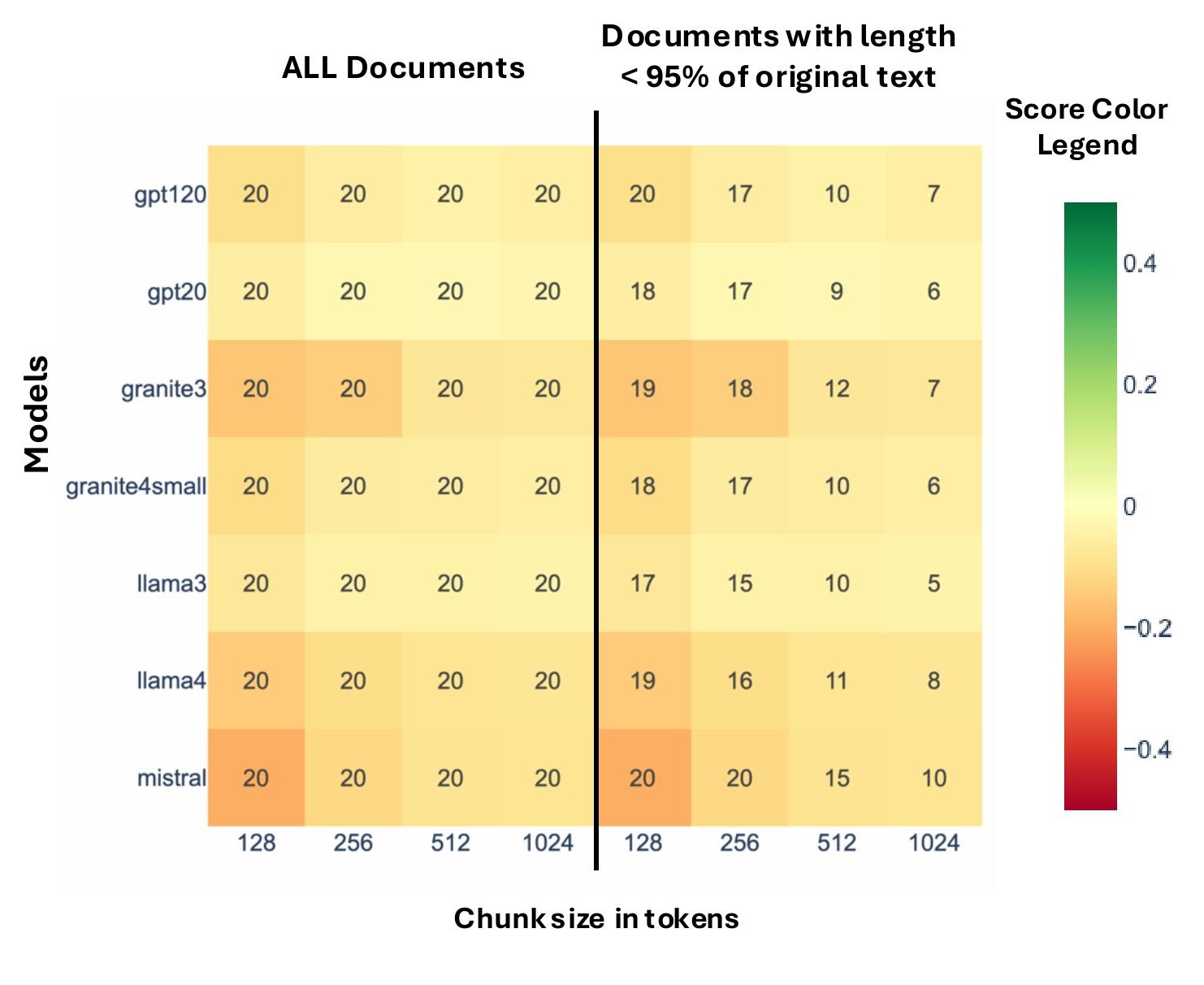}
    \caption{Provenance ROUGE-L recall scores relative to the full-text baseline after LLM condensing. Scores are averaged across fields and documents included in each panel. The left
    panel includes all 20 documents; the right includes only documents condensed to less than 95\% of their original
    length. Cell labels indicate document counts.}
    \label{fig:LLMRougeResults}
\end{figure*}

\begin{figure*}
    \centering
    \includegraphics[scale=0.2]{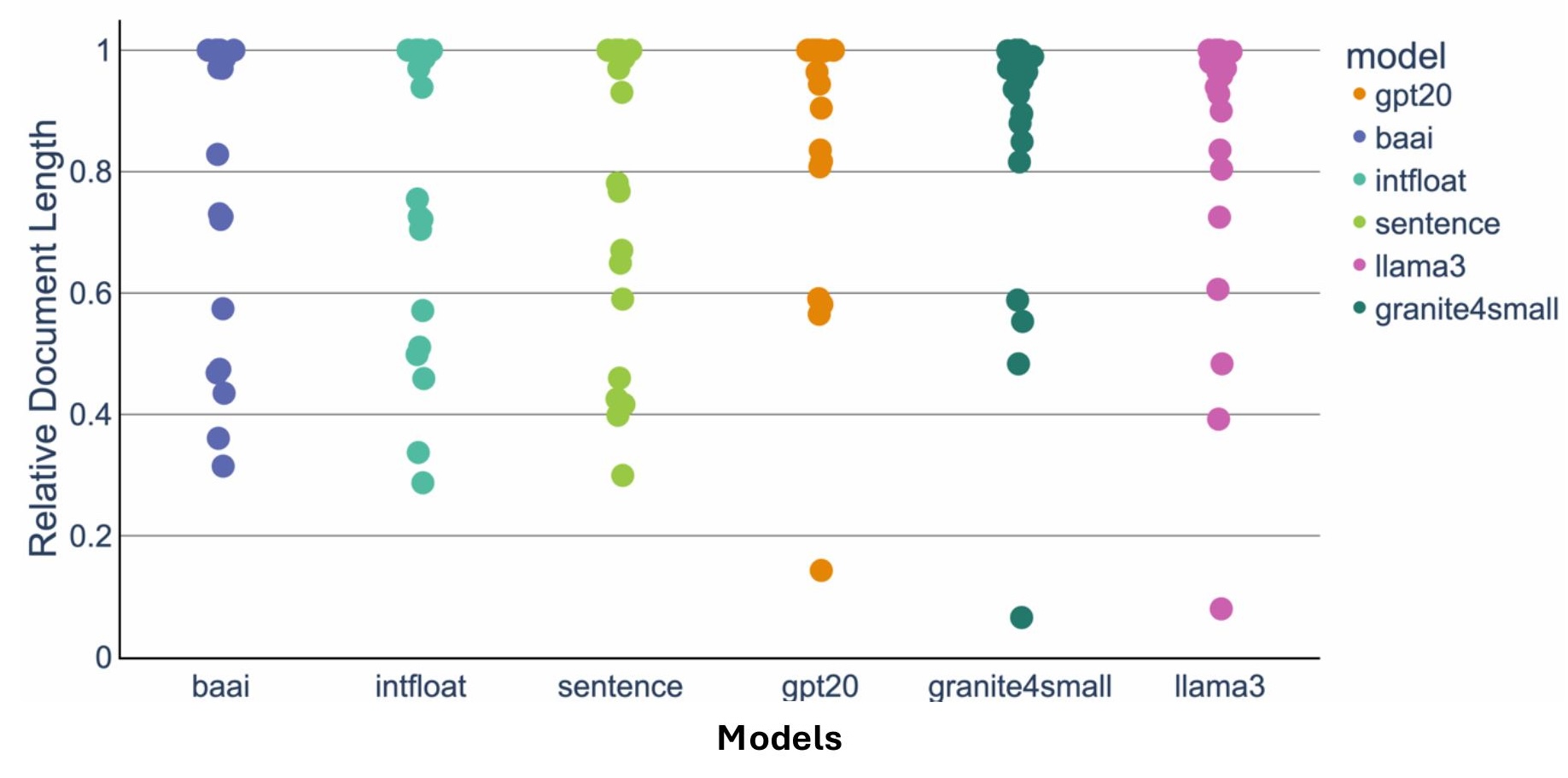}
    \caption{For each of the top performing models from the LLM protocol with chunk size of 512 tokens and RAG protocol with chunk size of 256 tokens and 10 chunks, each document length is represented by a point. The point is the percent of the document which is retained after the condensing protocol. }
    \label{fig:LengthScatter}
\end{figure*}

Figures~\ref{fig:RagRougeResults} and~\ref{fig:LLMRougeResults}
show provenance ROUGE-L recall relative to the full-text baseline
for RAG and LLM condensing, respectively.
Figure~\ref{fig:LengthScatter} shows the proportion of text
retained under the selected condensing configurations.

\begin{figure}[h!]
    \centering
    \includegraphics[width=\columnwidth]{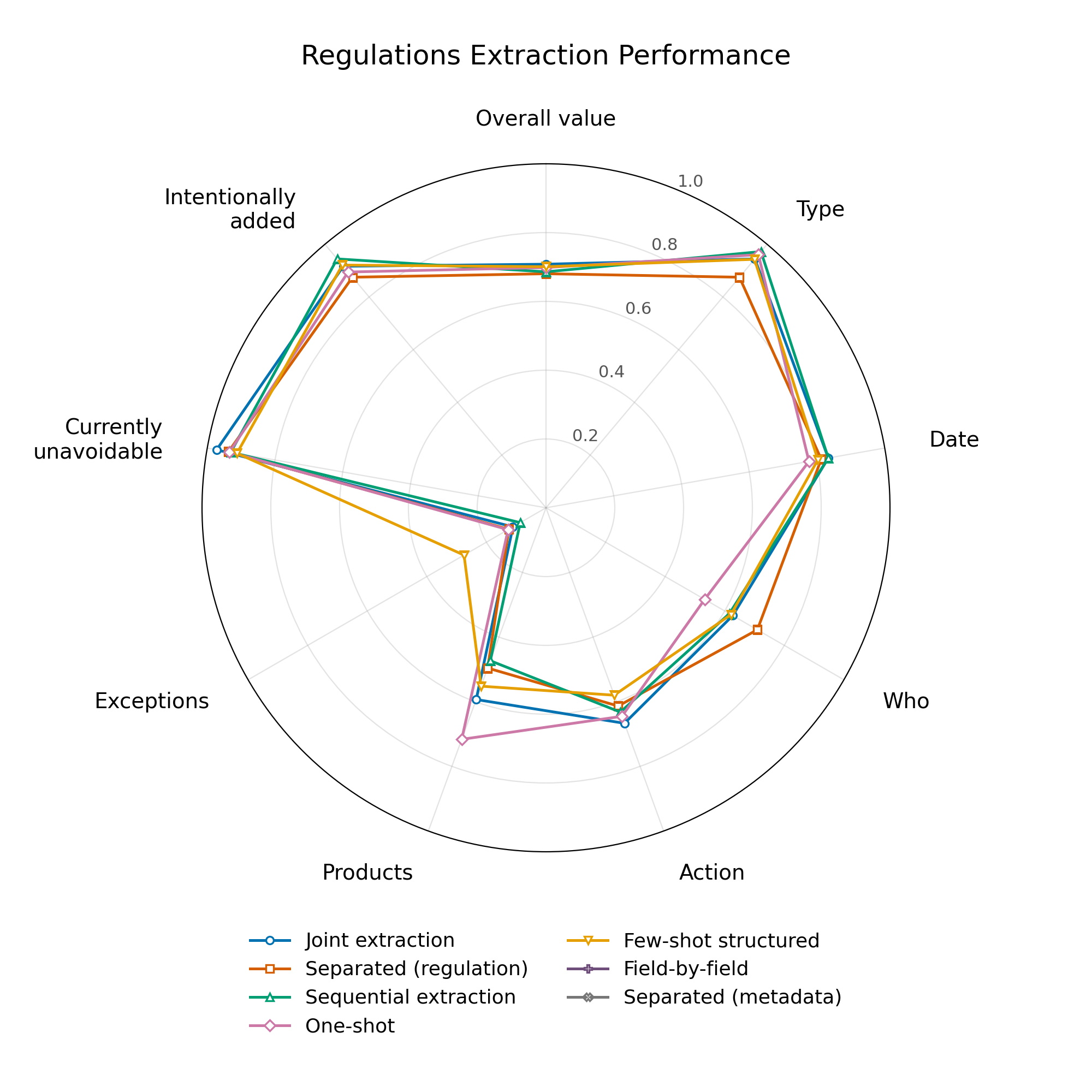}
    \caption{Regulation field evals after condensing with RAG}
    \label{fig:schema_regulation_radar_plot_rag}
\end{figure}

\begin{figure}[h!]
    \centering
    \includegraphics[width=\columnwidth]{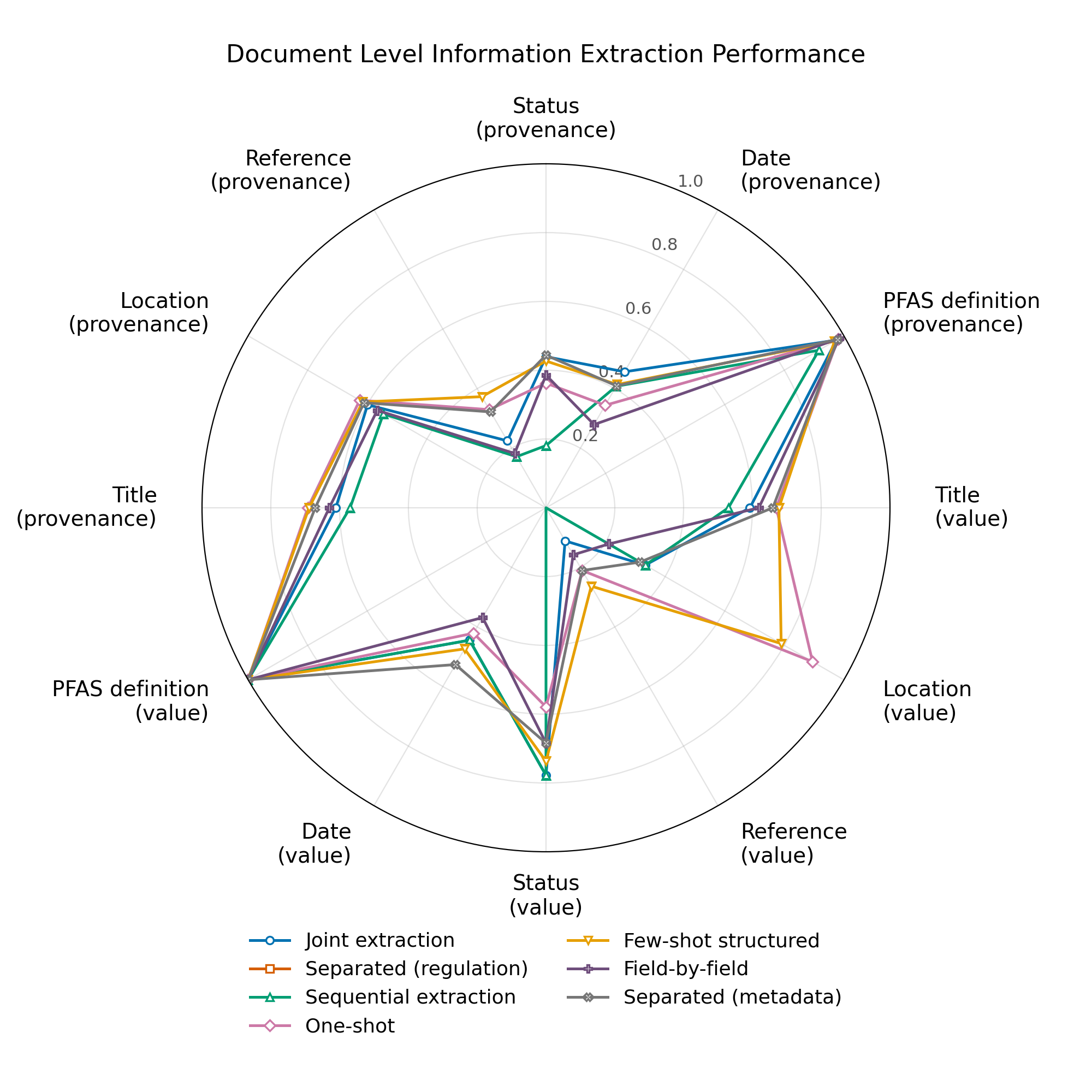}
    \caption{Document level field evals after condensing with RAG}
    \label{fig:schema_metadata_radar_plot_rag}
\end{figure}

\begin{figure}[h!]
    \centering
    \includegraphics[width=\columnwidth]{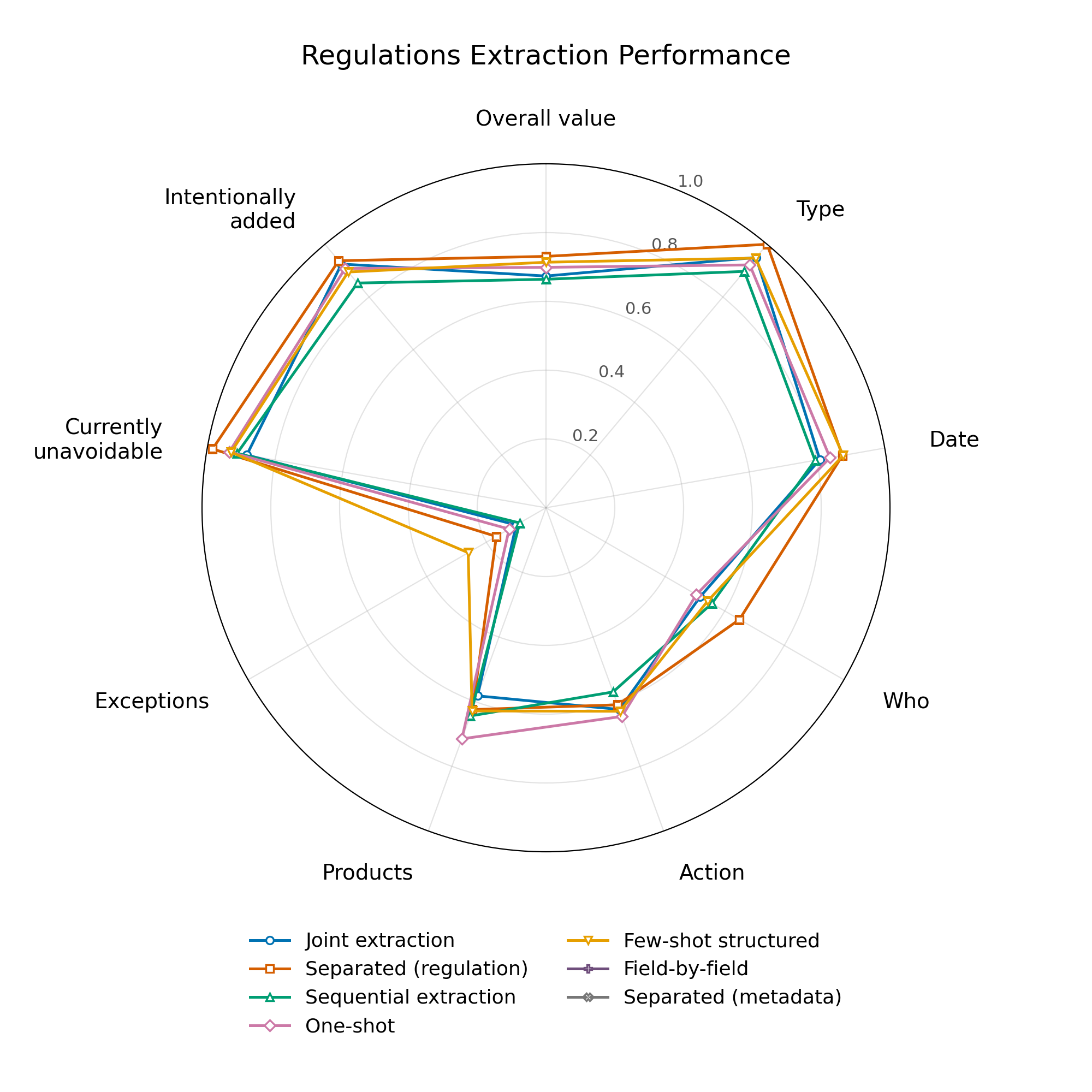}
    \caption{Regulation field evals after condensing with LLM}
    \label{fig:schema_regulation_radar_plot_llm}
\end{figure}

\begin{figure}[h!]
    \centering
    \includegraphics[width=\columnwidth]{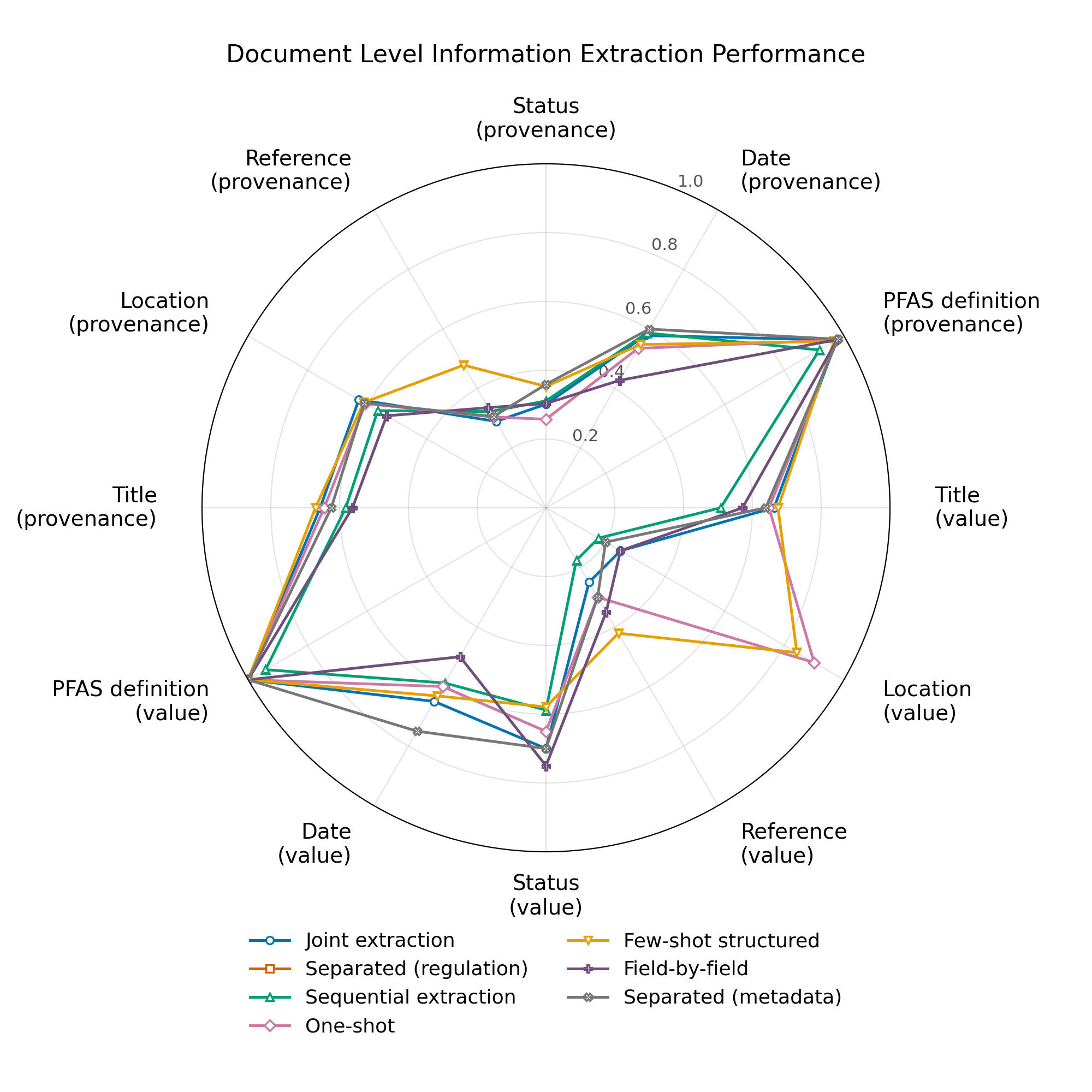}
    \caption{Document level field evals after condensing with LLM}
    \label{fig:schema_metadata_radar_plot_llm}
\end{figure}

The documents which were condensed with the top selected LLM or RAG processes were then used for information extraction. The results of these extractions are shown in figures \ref{fig:schema_regulation_radar_plot_rag} and \ref{fig:schema_metadata_radar_plot_rag} for RAG with 10 chunks and a chunk size of 256 tokens with sentence-transformers embeddings and figures \ref{fig:schema_regulation_radar_plot_llm} and \ref{fig:schema_metadata_radar_plot_llm} for LLM with a chunk size of 512 tokens with granite4.

\begin{table*}[t]
\caption{Comparison of GPT-4o prompting strategies on the RAG condensed text for PFAS regulation extraction. Metrics include regulation-level Precision (P), Recall (R), and $F_{1}$, along with averaged value and provenance scores for regulation fields.}
\centering
\small
\begin{tabular}{lccccc}
\hline
\textbf{Prompt Strategy } & \textbf{\#Valid}& \textbf{Reg. (P,R,$F_1$)} & \textbf{Reg. Value / Prov.} \\
\hline
Joint Extraction (baseline)      & 18&\textbf{1.00},0.84,0.88 & \textbf{0.71},0.61 \\
Separated (Regulation)   &16& 0.97,\textbf{0.89},0.90 & 0.68,0.61 \\
Sequential Extraction    & 18& \textbf{1.00},0.86,0.91 & 0.69,\textbf{0.62} \\
One-Shot                 & 19& 0.97,0.84,0.87 & 0.70,0.56 \\
Few-Shot Structured      & 19& \textbf{1.00},0.88,\textbf{0.92} & 0.70,0.60 \\
\hline
\end{tabular}
\label{tab:pfas_prompt_comparison_after_RAG}
\end{table*}

\begin{table*}[t]
\caption{Comparison of GPT-4o prompting strategies on the LLM condensed text for PFAS regulation extraction. Metrics include regulation-level Precision (P), Recall (R), and $F_{1}$, along with averaged value and provenance scores for regulation fields.}
\centering
\small
\begin{tabular}{lccccc}
\hline
\textbf{Prompt Strategy }  & \textbf{\#Valid}& \textbf{Reg. (P,R,$F_1$)} &\textbf{Reg. Value / Prov.} \\
\hline
Joint Extraction (baseline)      & 20& \textbf{1.00},0.78,0.85 & 0.67,0.61 \\
Separated (Regulation)   & 16 & 0.97,\textbf{0.93},\textbf{0.93} & \textbf{0.73},\textbf{0.62} \\
Sequential Extraction    & 17& \textbf{ 1.00},0.89,\textbf{0.93} & 0.66,0.62 \\
One-Shot                 & 20& 0.98,0.83,0.87 & 0.70,0.58 \\
Few-Shot Structured      & 19 & \textbf{1.00},0.83,0.88 & 0.71,0.58 \\
\hline
\end{tabular}
\label{tab:pfas_prompt_comparison_after_LLM}
\end{table*}

\newpage

\section{Prompting comparisons on condensed documents}
Comparisons of GPT-4o prompting strategies for regulation extraction from RAG- and LLM-condensed documents are presented in \autoref{tab:pfas_prompt_comparison_after_RAG} and
\autoref{tab:pfas_prompt_comparison_after_LLM}, respectively.

\section{Further information on benchmark construction}
The annotations were compiled by 2 domain experts who have experience in the field of PFAS regulations. The titles were chosen to be a descriptive phrase mentioned at the beginning of the document. This may or may not align with the header of the document. The location is where the regulation jurisdiction is, described by the location of the governing body. References are a shorthand for the document within the governing body's catalog, usually an alphanumeric combination. The document level date is the date that the document was written or submitted, it is unrelated to the date when the regulations become active. The pfas definition is outlined in the definitions section of each document, there are several commonly used definitions. The status of the document is always stated in the introduction, whether the following text is a proposal, a bill, or a law. Regulation information is almost always contained within single, separated paragraphs. The date, products, action, who, and intentionally added and currently unavoidable information should be within one paragraph for a single regulation. Exceptions are sometimes defined for the entire document as a whole outside of these paragraphs. Currently unavoidable and intentionally added fields are the most common qualifiers that were found in the documents, and are false unless they are mentioned in the text. Actions are shortened to the main verb phrases that are mentioned in the paragraph (e.g. "shall not be used for firefighting training purposes or testing purposes" $\rightarrow$ "used for firefighting training purposes" and "used for firefighting testing purposes"). 'Who' is the person or entity that the regulation is applied to. If there is no action named, it may be the verb form of the 'Who' (e.g. Who: Manufacturer $\rightarrow$ Action: manufacture). The date when the regulation becomes effective is almost always at the beginning of the regulatory sentence or paragraph. And the products should be the direct object of the regulation action. 

\end{document}